%% file: bmvc_review.tex
\documentclass{bmvc2k}

\usepackage[linesnumbered]{algorithm2e}
\usepackage{amsmath, amssymb}
\usepackage{tikz}
\usepackage{svg}
\usepackage{float}
\usepackage{cuted}
\usepackage{graphicx} 
\usepackage{multicol}

\usetikzlibrary{shapes.misc}
\tikzset{cross/.style={cross out, draw=black, ultra thick, minimum size=2*(#1-\pgflinewidth), inner sep=0pt, outer sep=0pt},
cross/.default={0.26cm}}

\newcommand{\sumlims}[2]{\sum\limits_{#1}\limits^{#2}}
\newcommand{\argmax}[2]{\argmax\limits_{#1}\left(#2\right)}
\newcommand{\logand}[2]{\bigwedge\limits_{#1}\limits^{#2}}
\newcommand{\logor}[2]{\bigvee\limits_{#1}\limits^{#2}}

\newtheorem{definition}{Definition}

\title{C-SWAP: Explainability-Aware Structured Pruning for Efficient Neural Networks Compression}

\addauthor{Baptiste Bauvin}{baptiste.bauvin@thalesgroup.com}{1}
\addauthor{Loïc Baret}{loic.baret@thalesgroup.com}{1}
\addauthor{Ola Ahmad}{ola.ahmad@thalesgroup.com}{1}

\addinstitution{
 \textbf{Thales}\\
 \textit{cortAIx Lab}\\
 Montreal, QC, Canada
}

\runninghead{B.Bauvin et al.}{C-SWAP: Explainability-Aware Structured Pruning}

\begin{document}
\addtolength{\tabcolsep}{-0.5em}

\maketitle

\begin{abstract}
Neural network compression has gained increasing attention in recent years, particularly in computer vision applications, where the need for model reduction is crucial to enable edge deployment constraints. Pruning is a widely used technique that prompts sparsity in model structures, e.g. weights, neurons, and layers, reducing size and inference costs. Structured pruning is especially important as it allows for the removal of entire structures, which further accelerates inference time and reduces memory overhead. However, it can be computationally expensive, requiring iterative retraining and optimization. To overcome this problem, recent methods considered one-shot setting, which applies pruning directly at post-training. Unfortunately, they often lead to a considerable drop in performance. In this paper, we focus on this issue by proposing a novel one-shot pruning framework that relies on explainable deep learning. 
First, we introduce a causal-aware pruning approach that leverages cause-effect relations between model predictions and structures in a progressive pruning process. It allows us to efficiently reduce the size of the network, ensuring that the removed structures do not deter the performance of the model.
Then, through experiments conducted on convolution neural network and vision transformer baselines, pre-trained on classification tasks, we demonstrate that our method consistently achieves substantial reductions in model size, with minimal impact on performance, and without the need for fine-tuning. Overall, our approach outperforms its counterparts, offering the best trade-off. Our code is available on https://github.com/ThalesGroup/C-SWAP.
\end{abstract}

\section{Introduction}
\label{sec:intro}
\begin{figure}[t]
    \centering
    \includegraphics[width=0.5\linewidth]{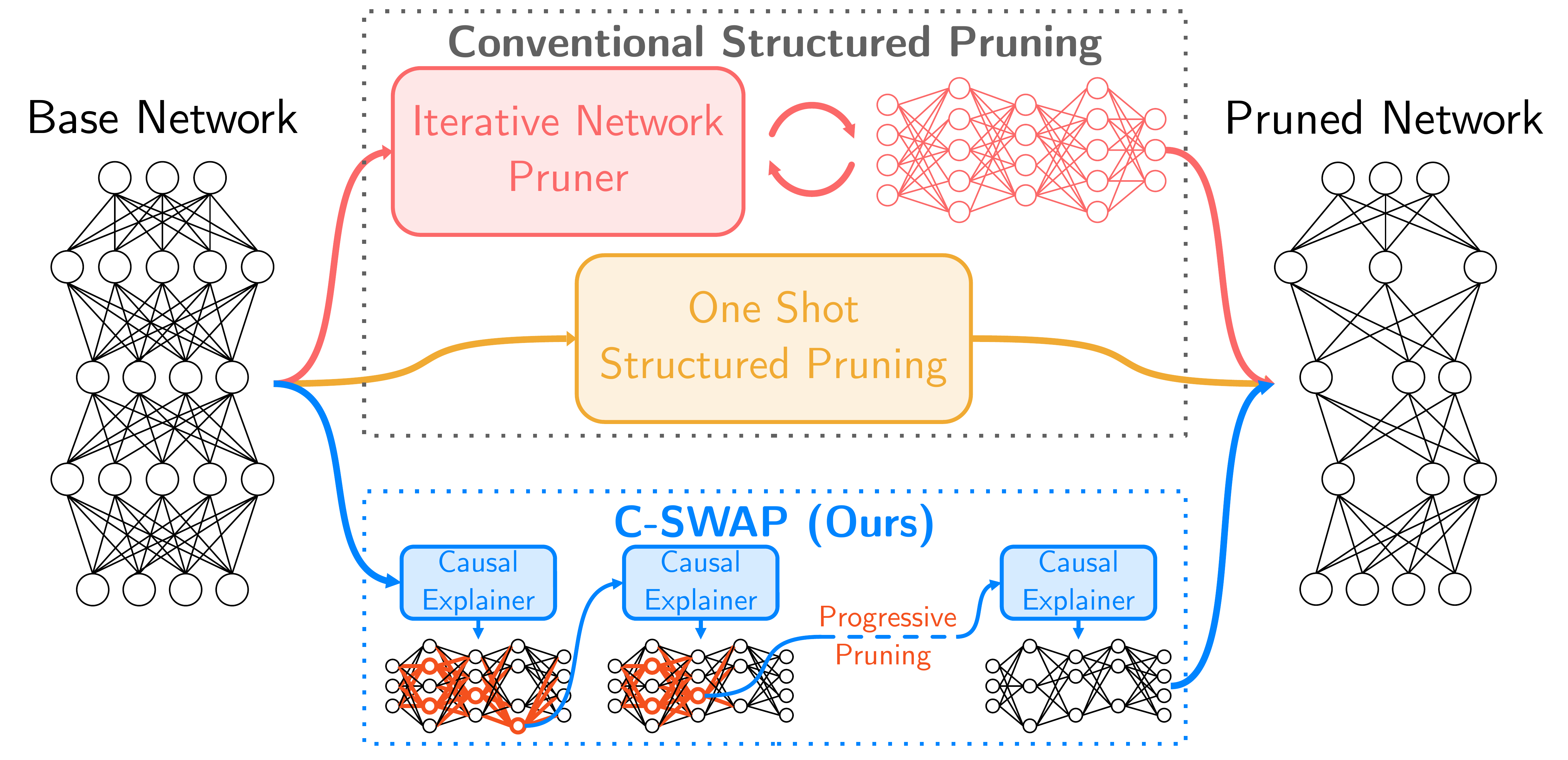}
    \caption{C-SWAP, an explainability-aware structured pruning framework for DNNs. Unlike conventional pipelines, C-SWAP uses a causal-based, one-shot progressive pruning strategy that removes neurons while preserving model performance without costly re-training. Its progressive approach scales effectively to complex architectures such as transformers.}
    \label{fig:teaser}
\end{figure}

Deep neural networks (DNNs) excel in computer vision tasks, yet they rely on computationally intensive training and inference processes \cite{reviewcnn}. This is often due to over- parameterization \cite{overparam}, where parameters are redundant, potentially degrading the model performance \cite{overparambad}. In addition, the large parameter footprints presents a significant challenge for deploying DNN models in resource-constrained environments. Hence, DNN compression techniques, particularly pruning \cite{Gao2021NetworkPV, Ding2020ResRepLC, Wang2020AccelerateCF, Yu2017NISPPN, OnlyTO, otov2, BiLevelOpt}, are vital for reducing model size, by removing redundant parameters \cite{Mao2017ExploringTG}, and accelerating the inference speed, while also supporting green machine learning (ML) by promoting sustainable practices \cite{greenpruning}.

Structured pruning (SP) effectively reduces DNN complexity by removing entire computational units, such as channels/filters, neurons, or layers, making it a practical solution for edge deployment \cite{OSSCAR, depgraph}. 
Typically, most SP methods need to be included into the training process \cite{OnlyTO, otov2, nmspar, chen2023overparameterized}, introducing additional computational overhead, instability during optimization, and sensitivity to hyper-parameters.  It can also involve a computationally intensive iterative process of pruning, and retraining, aiming to balance model sparsity with performance retention \cite{frankle2018lottery, iterPrun, Frankle2019LinearMC}. However, such a process poses scalability challenges when extended to complex architectures \cite{Zhang2021EfficientLT, OSSCAR} and conflicts with the green ML principles.

To overcome these limitations, this work focuses on post-training one-shot structured pruning (OSP), which removes whole structures after training, offering a more efficient alternative to iterative prune–retrain schemes \cite{SNIP, OSSCAR, ZipLM, SNIP, NetAF, PruningFF, OnlyTO}.
While it requires minimal fine-tuning, it introduces a notable trade-off between model complexity and performance, struggling with maintaining performance at high pruning thresholds \cite{BiLevelOpt}. 
Therefore, it is crucial to carefully select which parts of the model to prune to minimize fine-tuning while maintaining performance. For instance, one-shot magnitude pruning (OMP) \cite{magnitude} is widely used in computer vision due to its simplicity and applicability to structured pruning \cite{depgraph}. However, at higher pruning rates it can remove critical parameters, inducing performance regression. \textbf{\emph{This raises the question: can pruning be better guided by attribution signals from explainable AI (XAI), rather than magnitude alone?}}

More recently, researchers have explored XAI-guided pruning that leverages Layer-wise Relevance Propagation (LRP) attributions to rank units for removal and improve the reliability of the pruned models \cite{lrppruning}. While LRP-based approaches have shown promise when pruning is followed by post-pruning fine-tuning, our findings indicate that an alternative technique may enable more aggressive pruning ratios and substantially more compact models without additional fine-tuning. Nevertheless, most existing attribution methods were developed for comparatively simple CNNs and do not scale well to complex architectures (e.g., transformers) \cite{betterlrp}, or to dense-prediction tasks such as semantic segmentation.

Building on this motivation, we propose \textbf{C-SWAP}, a novel explainability-aware structured pruning method (see fig. \ref{fig:teaser}),  inspired by mechanistic interpretability research, which identifies prunable structures using causal inference.
Our method categorizes each neuron (channel in CNNs) as critical, neutral, or detrimental by perturbing their associated weights, computing the causal effect of these perturbations, and identifying significance using a statistical threshold. C-SWAP preserves critical neurons to maintain performance without post-pruning finetuning. To ensure C-SWAP's scalability for complex models, we couple our causal explanations with a progressive pruning process that scales to deep architectures such as large CNNs or vision transformers while maintaining stability. Our experiments show that C-SWAP demonstrates superior performance compared to various alternatives.
\newline
We summarize our \textbf{main contributions} as follows: (1) We first introduce a \textbf{multiclass, causal explanation criterion} to guide the pruning of classification models. (2) We present \textbf{C-SWAP}, a causal-guided pruning algorithm for deep and complex architectures that employs progressive pruning without fine-tuning. (3) Through experiments on CNNs and vision transformers on classification tasks, we show that our approach \textbf{outperforms all baseline pruning techniques} considered in this work. (4) We apply our approach to \textbf{semantic segmentation}, demonstrating its \textbf{extensibility to dense prediction tasks}.
\section{Related Work}
\label{sec:formatting}
\noindent\textbf{Pruning neural networks.}\,\,As DNNs become deeper and more complex, pruning is increasingly researched. Pruning falls into two categories: unstructured pruning \cite{whypruning}, which removes individual parameters, and structured pruning \cite{cheng2023survey}, which eliminates whole structures like filters, channels, or layers. Unstructured pruning has limited impact on achieving significant acceleration and compression for resource-constrained hardware \cite{OSSCAR, depgraph}. Our focus is structured pruning to enable efficient real-time computer vision applications.

\noindent\textbf{Structured pruning.}\,\,Structured pruning involves either one-shot pruning \cite{ZipLM, SNIP, NetAF, OSSCAR, onshot1} or iterative pruning \cite{frankle2018lottery, iterPrun, Frankle2019LinearMC, morcos2019one, zhou2019stabilizing}. Iterative pruning removes less significant components over multiple retraining cycles \cite{han2015learning, han2016deep}, exemplified by the Lottery Ticket Hypothesis (LTH) \cite{frankle2018lottery, lthstrucutred}. Though effective, iterative methods are computationally intensive \cite{morcos2019one, zhou2019stabilizing}. Our work aims at efficient OSP, applying post-training pruning without fine-tuning, reducing computational costs significantly \cite{OnlyTO}. Traditional one-shot techniques use weight magnitudes \cite{magnitude}, neuron relevance \cite{actrel}, or second-order derivatives \cite{lecun} for pruning, but recent work suggests these are suboptimal compared to criteria derived from XAI \cite{lrp1, magbad}.

\noindent\textbf{Explainability for OSP.}\,\,Explainability in pruning has been initially investigated with Layer-Wise Relevance Propagation (LRP) \cite{lrppruning, lrp1} that attributes importance scores to neural network internal structures. Initially applied to neuron masking \cite{lrppruning, betterlrp, lrp4}, it does not capture the complex interactions necessary for physically removing neurons without compromising the architecture, particularly in models with residual connections. Similarly, Amortized Explanation Methods \cite{aem} use trained networks to predict saliency maps to guide the pruning. They are effective but restricted to CNNs and require training an additional network. DeepLIFT \cite{deeplift} has also been used for filter-level pruning \cite{deepliftpruning} but is less effective due to noise sensitivity and 

\noindent\textbf{Progressive pruning for model explainability.}\,\, Mechanistic interpretability uses pruning to identify critical sub-structures via mask learning \cite{modular, subnetprobing} or progressive pruning, exemplified by Automatic Circuit DisCovery (ACDC) \cite{acdc}. Though scalable to large architectures, these methods are aggressive (see sec. \ref{sec:benchmark}) and tailored for task-specific explainability. Our work combines explainability for pruning and pruning for explainability by integrating a robust attribution method with progressive pruning in a unified DNN compression algorithm.

\begin{figure*}[t]
    \centering
    \begin{minipage}{0.49\linewidth}
    \centering
        \includegraphics[width=0.6\linewidth]{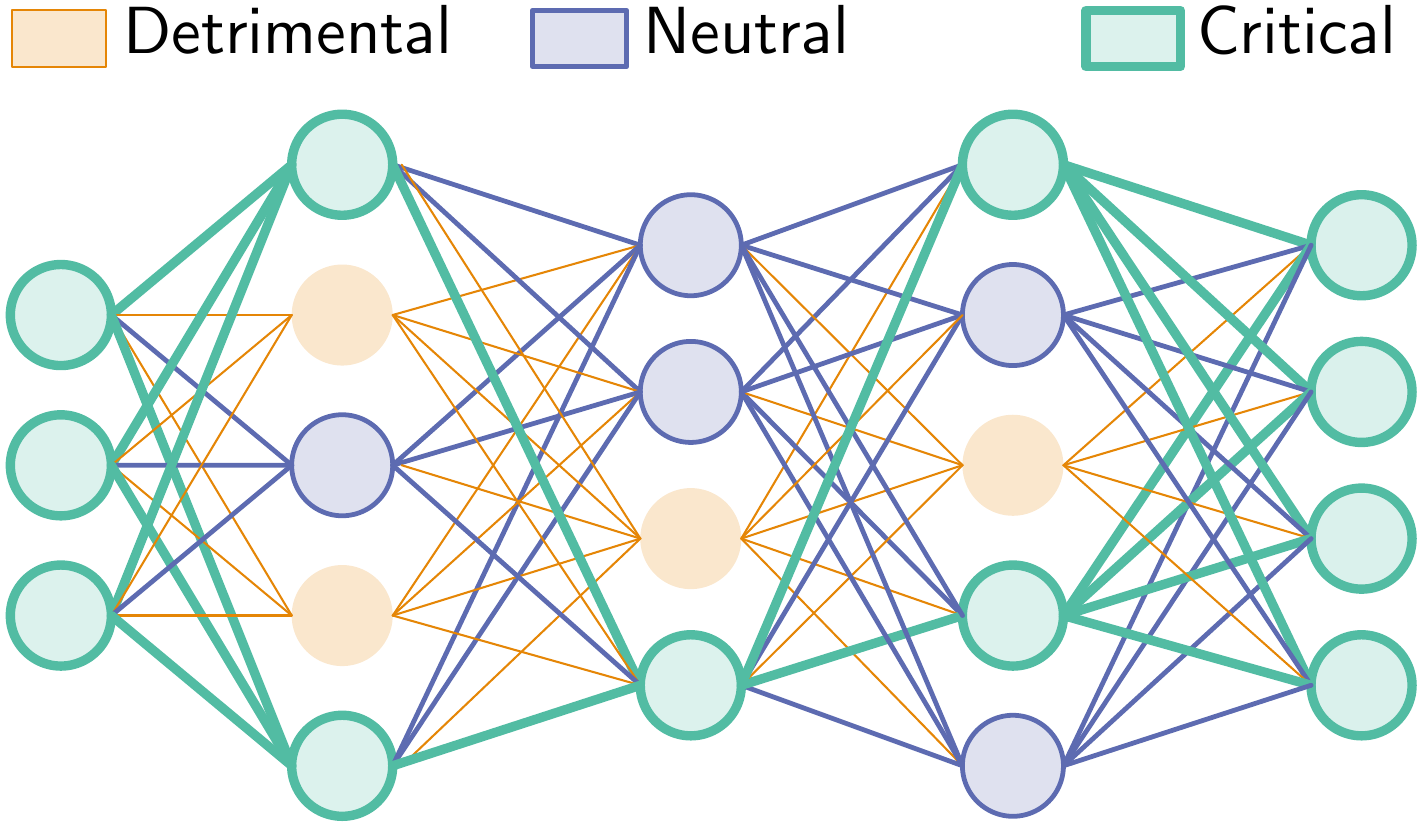}
\caption{Neurons and paths for a toy MLP. Critical path conveys the relevant information. Detrimental and Neutral neurons are not essential.}
\label{fig:causal_net}
    \end{minipage}%
    \hfill
    \begin{minipage}{0.49\textwidth}
        \centering
\input{mlp_res}
\caption{Simple MLP with a residual connection (dotted). Structured pruning forces to remove all the weights in red to remove the crossed neuron.}
\label{fig:pruning}
    \end{minipage}%
\end{figure*}

\section{Method}

We propose C-SWAP, a novel causal-aware structured pruning algorithm that consolidates a generalized causality-based explanation with progressive pruning techniques. Drawing inspiration from recent works \cite{ahmad2024causal, acdc}, C-SWAP leverages a pre-trained network alongside a set of class examples to compute a generalized causal effect that unveils the underlying causal mechanisms of model prediction (sec. \ref{sec:gce}). This causal effect is evaluated with a statistical significance test, allowing the classification of neurons in three categories: critical, detrimental, and neutral (sec. \ref{sec:gce} \& fig. \ref{fig:causal_net}).
Guided by this classification, C-SWAP structurally prunes the neutral and detrimental neurons while preserving the critical ones, processing the network from the output layer back to the input one. Structured pruning is shown in fig. \ref{fig:pruning}.
\subsection{Notations}
\label{sec:notations}
We consider $x$ to be the input of the network $F$, and we denote by $y$ its label with $y \in \{1, .., C\}$ ($C$ being the total number of classes). The output logits of $F$ given input $x$ are denoted $\{z_k\}_{k=1}^C$. Consequently, we denote by $\hat{y}$ the prediction of $F$ for $x$ defined as $\hat{y} = \arg \max_{k \in \{1,..,C\}}(z_k)$. Furthermore, we denote by $\hat{p}_k$ the probability of predicting class $k$ by $F$.
We consider that $F$ is composed of $L$ layers indexed by $l$, each containing $N_l$ neurons (channels in CNNs), denoted by their index $n_l$. The total number of neurons in the network is denoted $N = \sum_{l=1}^{L}N_l$.
Finally, we consider a set $\mathcal{S}$ of $M$ samples, partitioned in $C$ class-specific sets $\{\mathcal{S}_k\}_{k=1}^C$ of sizes $M_k$.

\subsection{Multi-class causal inference criterion} 
\label{sec:gce}
The core concept of class-specific causal effect, as introduced in \cite{ahmad2024causal}, involves performing path interventions to evaluate the importance of individual neurons.
This is achieved by removing a subset of weights out-coming from a neuron $n_{l}$ at layer $l$, and then measuring the resulting change (or divergence) in the model’s prediction distribution for a specific class $k$ across examples $x_{k}\in \{1,...,M_{k}\}$.
In our work, we extend this approach to multi-class classification, enabling us to compute the global effect of an intervention. This generalized causal effect not only captures inter-class dependencies but also serves as a robust criterion in our novel pruning framework. 
\begin{definition}[Global Causal Effect]
    Given a scoring function $\sigma$, \textbf{our global causal effect} of neuron $n$, its causal effect on samples from $C$ classes is defined as 
    $\xi_n = \frac{1}{M}\sumlims{x \in \mathcal{S}}{}\frac{\sigma^*_{n}(x) - \sigma(x)}{\sigma(x)}, $
    \label{def:gce}
\end{definition}
where $\sigma^{*}_{n}(x)$ and $\sigma_{n}(x)$ are scoring functions respectively derived from the perturbed and the original networks $F^{*}$, $F$, given an input $x$. The scoring function can be defined by any metric that characterizes model performance. In the classification setting, we derive it from the true class probability $\sigma_{\text{Classif.}}:  (x, y) \rightarrow \exp(z_y)/\sum_{k=1}^{C}\exp(z_k) $.

The global causal effect is a signed measure that quantifies a neuron's overall influence and potential role: a positive value ($\xi_n > 0$) indicates a deleterious impact, and a negative one ($\xi_n < 0$) a beneficial impact.

We assess each neuron's influence by comparing the distribution of the perturbed scoring function $\sigma_{n}^{*}(x)$ with the initial one $\sigma_{n}(x)$. Significant divergence between these distributions indicates a neuron's impact on model performance.
To infer significant divergence, we use hypothesis testing for each class-specific subset of samples $\mathcal{S}_k$, computing a predicate $\pi^{(k)}_n$. This predicate is true if the neuron's effect is statistically significant at a 5\% level for class $k$, and false otherwise.

In practice, we apply a paired t-test to compare $\sigma_{n}$ and $\sigma^{*}_{n}$ for each class $k$ and neuron $n$, estimating $\pi^{(k)}_n$ to quantify the effect of each neuron across classes.
Unlike the class-specific method \cite{ahmad2024causal}, which directly applies a statistical threshold to analyze the neurons, we consider a \textit{voting} strategy (see def. \ref{def:crit}) to address the complexity of the multi-class problem.
\begin{definition}[Neutral, Critical and Detrimental neurons]
A neuron $n$  is considered \textbf{Neutral} if $\logand{k=1}{C} \neg\, \pi_n^{(k)},$
\textbf{Critical} if $\left(\logor{k=1}{C} \pi_n^{(k)}\right) \wedge  (\xi_n < 0$), and
\textbf{Detrimental} if $\left(\logor{k=1}{C} \pi_n^{(k)}\right) \wedge (\xi_n > 0$). With $\neg$, the logical negation sign.
\label{def:crit}
\end{definition}
A neuron $n$ is \textit{neutral} if its influence is statistically insignificant for all classes. It is \textit{critical} if it significantly affects at least one class and enhances predictions. Conversely, a neuron is \textit{detrimental} if it significantly affects at least one class but degrades model performance.

To summarize, we perform class-specific statistical inference and aggregate results using the voting scheme from def. \ref{def:crit}. 
To safely prune neurons, we consider each $\pi_n^{(k)}$ independently, avoiding the removal of class-specific information, to the cost of potentially more false positives. Analyzing significance across all samples of $\mathcal{S}$ might imply missing neurons impacting a single class, as discussed in sec. \ref{sec:ablation}.
\subsection{C-SWAP}
\label{sec:ranking}
\newcommand{\algsize}{0.8}
\SetKwComment{Comment}{\#\,}{}
\RestyleAlgo{ruled}
\SetAlgoNoEnd
\begin{algorithm}[t]
\label{alg:cswapp}
\scriptsize
\caption{C-SWAP algorithm.}
\KwData{Pre-trained DNN: $F$ ; Manifold $\mathcal{S}$ ; Scoring function: $\sigma(x)$; P-Value: $\alpha = 0.05$}
$G \gets F$ \Comment*[r]{\scalebox{\algsize}{initialize network}}
\For{layer $l$ in $[L-1, .., 1]$ (bottom up)}{
  \For{neuron $n$ in layer $l$}{
    $\Tilde{G} \gets G$ cutting $\{n \rightarrow \nu \}$ for $\nu \in l+1$ \Comment*[r]{\scalebox{\algsize}{connections to neurons of next layer}}
  Compute $\xi_n$  \Comment*[r]{\scalebox{\algsize}{general causal effect}}
  \For{class $k$ in $\{1, C\}$}{
        Compute $\pi_n^{(k)}$ \Comment*[r]{\scalebox{\algsize}{statistical inference}}
    }
  \If{$\logor{k=1}{C} \pi_n^{(k)}$ and $\xi_n \leq 0$}{$\mathcal{C} \gets \mathcal{C} \cup \{n\} $ \Comment*[r]{\scalebox{\algsize}{neuron is critical} }}
  \Else{$G \gets \Tilde{G}$  \Comment*[r]{\scalebox{\algsize}{neuron is not critical: remove it}}}}}
\For{neuron $n$ in $\mathcal{C}$ (ranked by $\xi_n$)}{Prune $n$ from $G$ \Comment*[r]{\scalebox{\algsize}{if needed remove critical neur.}}}
\KwResult{Pruned network $G$}
\end{algorithm}
The primary goal of pruning is to remove as many neurons as possible without compromising performance. Traditional XAI-based OSP strategies such as \cite{lrppruning, deepliftpruning, betterlrp} face challenges when pruning large portions of the network's neurons without fine-tuning. These strategies involve analyzing the pre-trained network, ranking neurons based on their importance criterion, and subsequently removing the least relevant neurons. While effective for small-scale pruning, the ranking quality of these methods diminishes significantly when pruning a large number of neurons (see fig. \ref{fig:short}). This occurs because the ranking is computed globally over the entire network before pruning.

Ideally, to preserve the performance as long as possible, one would remove the least important neuron, recompute the ranking on the pruned network, and then iterate as many times as necessary, removing one neuron and re-ranking. This iterative process, similar to a greedy algorithm, ensures nearly optimal OSP. However, this approach is computationally prohibitive for DNNs.

To address this limitation, we propose C-SWAP, that leverages properties of our causal-based criterion to relax the greedy algorithm. Instead of identifying and removing the absolute least important neuron at each iteration, C-SWAP captures neurons that are not critical to the network's functionality, without the necessity for a global ranking. As a consequence, when an ideal greedy algorithm removing $m$ neurons would require a costly ($\mathcal{O}(N)$) analysis of the DNN for each neuron pruned; hence a $\mathcal{O}(m \times n)$ total complexity, C-SWAP runs through the process in $\mathcal{O}(n)$. Indeed, it allows to efficiently iterate over all neurons, systematically pruning irrelevant neurons during the course of the analysis, ensuring scalability for DNNs while maintaining the performance through the pruning process. Consequently, C-SWAP is not more computationally expensive than any ranking-based XAI pruning method, as it intertwines the analysis and pruning processes.

Technically, C-SWAP, presented in alg. \ref{alg:cswapp}, integrates the intervention strategy introduced in sec. \ref{sec:gce} in the pruning process. For each neuron, we compute its general causal effect and predicates $\pi^{(k)}_n$. Then, if the neuron is not deemed critical by the causal analysis, it is systematically removed from the network. If the neuron is deemed critical, it is ranked among all other critical ones for pruning at the end of the process, if required. This strategy leverages the advantages of the optimal greedy method without any computational overhead compared to traditional XAI rankings, thanks to the causal explanations that easily allow to detect critical neurons without the need for a global ranking.
\begin{table}[t]
        \centering
        \scriptsize
        \begin{tabular}{|l|p{2.2cm}|p{8cm}|}
        \hline
        \textbf{Method} & \textbf{Principle} & \textbf{Description} \\
        \hline
        Integ. Grad. \cite{integratedgradients} & Baseline comparison & Averages gradients from a baseline to actual input, providing smooth feature importance attribution. \\
        \hline
        DeepLIFT \cite{deeplift} & Reference comparison & Compares activations between actual input and baseline to quantify contributions, offering stability over gradients alone. \\
        \hline
        iLRP \cite{betterlrp} & Relevance allocation & Allocates relevance scores to neurons, distributing prediction score across layers to highlight important contributions. \\
        \hline
        Intern. Infl. \cite{internalinfluence} & Internal role clarification & Details the contributions of internal neurons/layers to model output for specific input. \\
        \hline
        Conductance \cite{conductance} & Gradient flow & Assesses influence of neurons/layers by analyzing gradient information flow, similar to Integrated Gradients. \\
        \hline
        AMP \cite{acdc} & Circuit Discovery & Finds the optimal sub-circuit of a DNN for a specific task by gradually pruning its non-relevant components. \\
        \hline
    \end{tabular}
    \caption{Summary of existing explanation methods used in our classification experiments.}
    \label{tab:explanation_criteria}
\end{table}
\subsection{Key parameters of the methods}
Our framework depends on a set of parameters, beginning with a scoring function $\sigma$ (def. \ref{def:gce}) that we define as the probability of correct classification $\hat{p}_y$.
Second, it employs a hypothesis test evaluated over a data manifold $\mathcal{S}$, consisting of $M$ inputs, which we set to $M_k = 128$ samples by class. In app. \ref{app:samples}, we show that increasing it has minimal impact on C-SWAP and that lower amount of samples still produce reliable causal inference. With such a small sample budget, C-SWAP can operate on a rebalanced dataset, thereby compensating for class imbalance in the original dataset. Finally, C-SWAP relies on a significance level $\alpha$, which we set to $5\%$ and explore its impact in sec. \ref{sec:ablation}.
\section{Experiments}
\label{sec:exps}
\label{sec:models_data}
\renewcommand{\arraystretch}{1.2}
\begin{figure*}[t]
    \centering
    \begin{minipage}{0.48\linewidth}
        \centering
        \begin{table}[H]
\scriptsize
    \vspace{-3mm}
    \centering
    \begin{tabular}{|l|c|c|c|c|c|c}
    \hline
        \textbf{Model} & \textbf{ResNet-18} & \textbf{MobileNetV2} & \textbf{ResNet-50} & \textbf{ViT}  \\
    \hline
        \# Params (M) & 11.4 & 3.9 & 23.9 & 86 \\
        \hline
    \# Layers & 18 & 28 & 50 & 24 \\
    \hline \hline
    \textbf{Dataset} & \textbf{CIFAR10} & \multicolumn{3}{c|}{\textbf{ImageNet}} \\
    \hline
        \# Sample/class & 6000 & \multicolumn{3}{c|}{400} \\
    \hline
    Image size & 32x32x3 & \multicolumn{3}{c|}{224x224x3}\\
    \hline
    \end{tabular}
    \vspace{12mm}
    \caption{Summary of models and datasets used in our experiments.}
    \label{tab:my_label}
\end{table}
    \end{minipage}%
    \hfill
    \begin{minipage}{0.45\textwidth}
    \vspace{-4mm}
        \centering
        \includegraphics[width=0.56\linewidth]{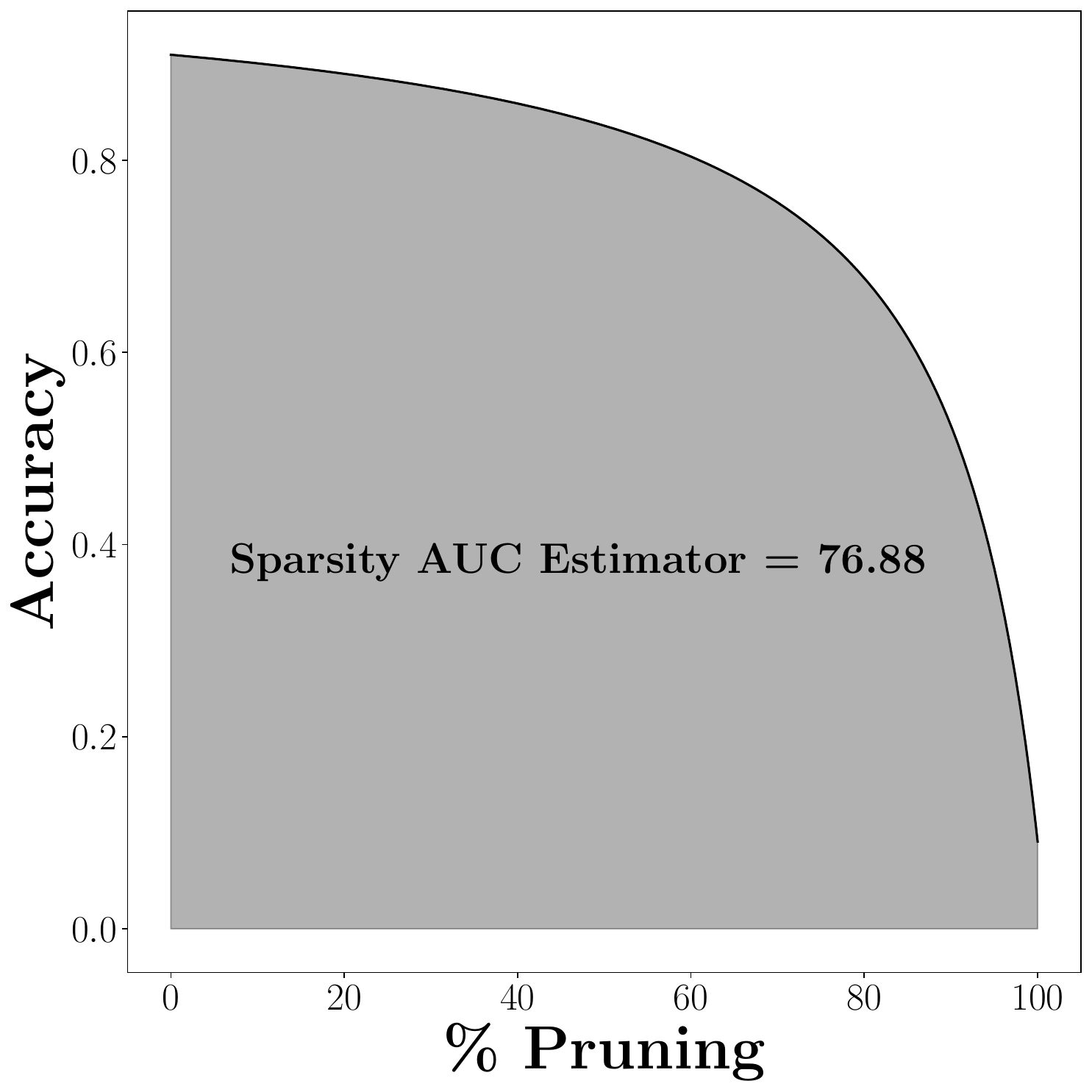}
    \caption{SAUCE for a pruning curve.}
    \label{fig:sauce}
    \end{minipage}%
\end{figure*}
\renewcommand{\arraystretch}{1}
Tab. \ref{tab:my_label} summarizes the models and datasets used in our experiments. ResNet-18 \cite{resnet18}, represents a medium-sized architecture, while ResNet-50 \cite{resnet18}, exemplifies a larger network. MobileNetV2 \cite{mobilenet}, serves as a compact, high-density model. Finally we include ViT \cite{vit}, a vision transformer network. For more information on pruning ViT see app. \ref{app:struct}.
We selected two datasets of increasing difficulty. CIFAR10 \cite{cifar10} consists of small resolution images, and is widely used to evaluate and benchmark compression methods in classification architectures. For a more complex task, we include a subset of ImageNet \cite{imagenet} consisting of 10 classes of interest, as reported in \cite{ahmad2024causal}.
\subsection{Baselines, implementations and evaluation}
\begin{figure}
\newcommand{\mysize}{0.24}
\scriptsize
\begin{tabular}{cccc}
 \multicolumn{4}{c}{\includegraphics[width=0.95\textwidth, clip, trim=0.2cm 0.4cm 0.2cm 1.85cm,]{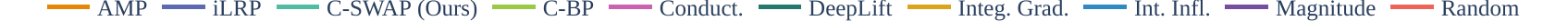}}\\
 \includegraphics[width=\mysize\linewidth]{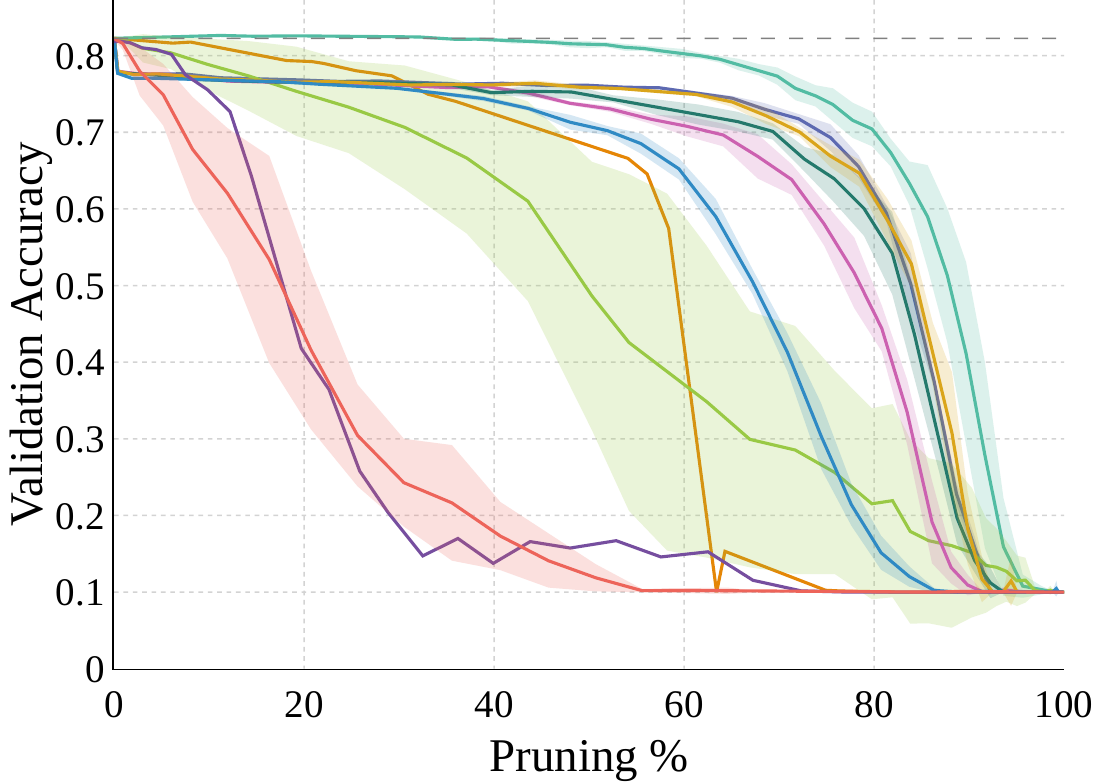}&
 \includegraphics[width=\mysize\linewidth]{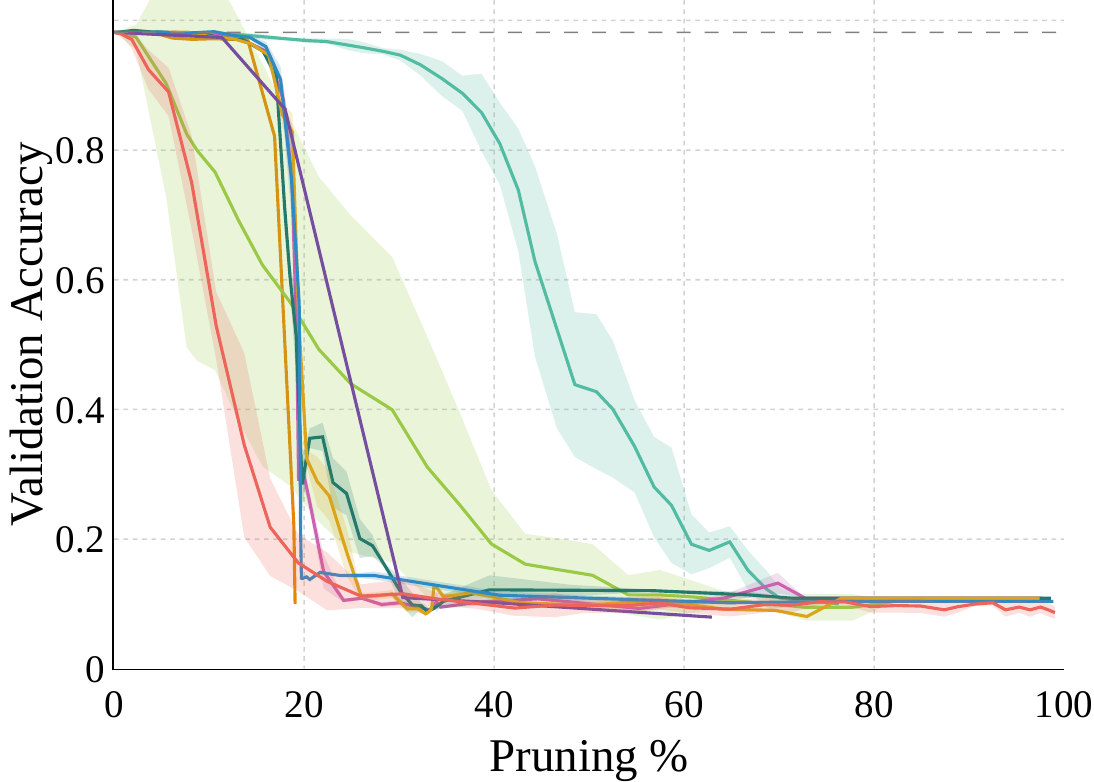}&
 \includegraphics[width=\mysize\linewidth]{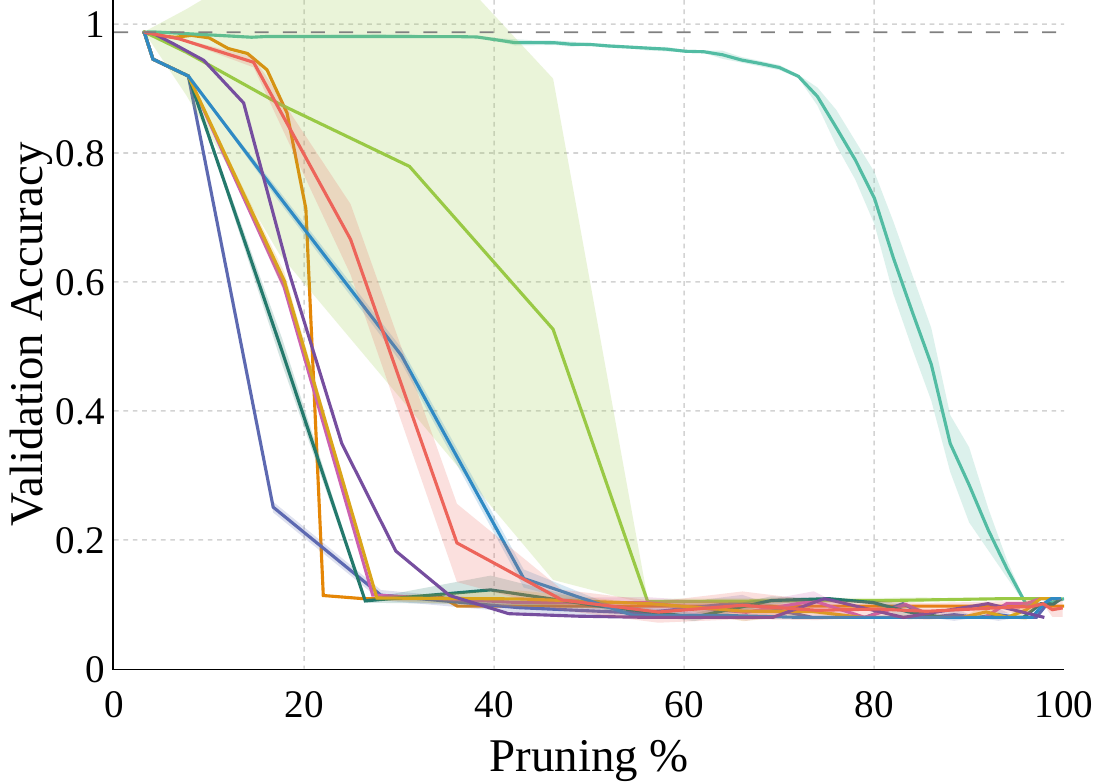} &\includegraphics[width=\mysize\linewidth]{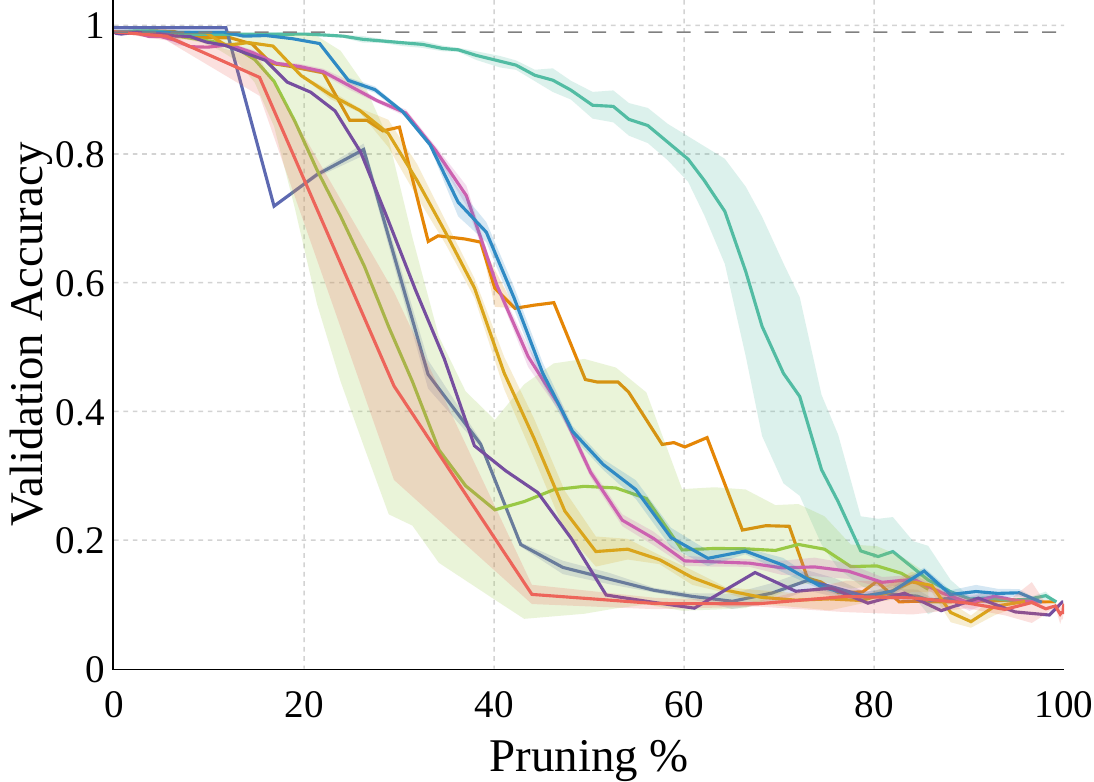}\\
 (a) ResNet-18 on CIFAR10.& (b) MobileNetV2 on ImageNet.& (c) ResNet-50 on ImageNet.&(d) ViT on ImageNet.
\end{tabular}
\caption{Validation accuracy as a function of percentage of parameters removed for four architectures. Dashed line represents original model performance. }
\label{fig:short}
\end{figure}

We use two baseline pruning methods: random pruning \cite{random} and OMP \cite{magnitude}, a gold standard for evaluating state-of-the-art pruning techniques.
We add various explanation approaches reported in tab. \ref{tab:explanation_criteria} and include a pruning method derived from ACDC \cite{acdc}, that we denote "Adapted Mechanistic Pruning" (AMP), as detailed in app. \ref{app:acdc}. Finally, we include C-BP (see app. \ref{app:cbp}) a version of C-SWAP that does not include the progressive pruning process, but simply prunes the least important neurons, according to the causal criterion, to highlight the importance of the progressive pruning strategy. Implementations utilize Captum \cite{captum} and DepGraph \cite{depgraph} libraries or authors' code. 
To evaluate pruning impact, we propose two assessment methods. The pruning curve that computes average accuracy over the validation set as a function of pruning percentage (fig. \ref{fig:short}). And the novel Sparsity AUC Estimator (SAUCE) that quantifies the area under the accuracy curve (fig. \ref{fig:sauce}). SAUCE provides a single value assessing each pruning criterion's effectiveness in preserving information across pruning ratios.
Evaluation metrics are averaged over five random seeds, reported in relation to pruned parameter percentages, adhering to structured pruning literature. App. \ref{fig:compression} shows the quasi-linear correlation between pruning percentage and size \& FLOPs reduction, making it a good proxy for compression. In addition app. \ref{app:time} shows the computation time of C-SWAP.

\renewcommand{\arraystretch}{1.2}
    \begin{table}[t]
        \centering
        \small
\begin{tabular}{l|c|c|c|c||c}
\hline
 & ResNet18 (C10) & ResNet50 (IN) & MobileNet (IN) & ViT (IN) & Aver. \\
\hline
Random & $25.0{\scriptstyle\,\pm\,2.13}$ & $30.8{\scriptstyle\,\pm\,1.28}$ & $19.8{\scriptstyle\,\pm\,0.81}$ & $34.2{\scriptstyle\,\pm\,1.72}$ & 27.4 \\
OMP \cite{magnitude} & $26.0{\scriptstyle\,\pm\,0.0}$ & $24.0{\scriptstyle\,\pm\,0.0}$ & $26.3{\scriptstyle\,\pm\,0.0}$ & $40.1{\scriptstyle\,\pm\,0.0}$  & 29.1 \\
AMP & $49.6{\scriptstyle\,\pm\,0.0}$ & $24.6{\scriptstyle\,\pm\,0.0}$ & $17.3{\scriptstyle\,\pm\,0.0}$ & $42.7{\scriptstyle\,\pm\,8.16}$ &  33.6 \\
DeepLIFT \cite{deeplift}& $64.1{\scriptstyle\,\pm\,0.67}$ & $21.2{\scriptstyle\,\pm\,0.19}$ & $28.6{\scriptstyle\,\pm\,0.25}$ & - &  37.9 \\
C-BP (ours) & $49.5{\scriptstyle\,\pm\,5.44}$ & $42.3{\scriptstyle\,\pm\,9.73}$ & $28.7{\scriptstyle\,\pm\,2.43}$ & $38.5{\scriptstyle\,\pm\,5.19}$ & 39.7 \\
Conductance \cite{conductance} & $61.5{\scriptstyle\,\pm\,0.61}$ & $22.4{\scriptstyle\,\pm\,0.26}$ & $27.1{\scriptstyle\,\pm\,0.07}$ & $48.1{\scriptstyle\,\pm\,0.13}$ & 39.8 \\
Integ. Grad.  \cite{integratedgradients} & $65.5{\scriptstyle\,\pm\,0.14}$ & $22.4{\scriptstyle\,\pm\,0.26}$ & $27.6{\scriptstyle\,\pm\,0.15}$ & $44.0{\scriptstyle\,\pm\,0.35}$ &  39.9 \\
iLRP \cite{betterlrp} & $65.6{\scriptstyle\,\pm\,0.18}$ & $18.2{\scriptstyle\,\pm\,0.18}$ & - & $36.9{\scriptstyle\,\pm\,0.25}$&  40.3 \\
Inter. Infl. \cite{internalinfluence} & $55.2{\scriptstyle\,\pm\,0.45}$ & $29.1{\scriptstyle\,\pm\,0.23}$ & $27.6{\scriptstyle\,\pm\,0.09}$ & $49.3{\scriptstyle\,\pm\,0.09}$& 40.3 \\ \hline\hline
\textbf{C-SWAP (Ours)} & $\mathbf{72.1{\scriptstyle\,\pm\,0.39}}$ & $\mathbf{80.3{\scriptstyle\,\pm\,0.28}}$ & $\mathbf{49.4{\scriptstyle\,\pm\,1.15}}$ & $\mathbf{68.1{\scriptstyle\,\pm\,1.67}}$& $\mathbf{67.5}$ \\
\hline
\end{tabular}
        \caption{\textbf{SAUCE} measuring the strength of each pruning criterion.}
\label{tab:acc}
\end{table}

\renewcommand{\arraystretch}{1.1}

\subsection{Comparative results}
\label{sec:benchmark}
We evaluate the impact of explanation methods on model pruning by calculating average accuracy at increasing pruning ratios, without fine-tuning.
Fig. \ref{fig:short}.a shows ResNet-18 on CIFAR10, where most explanation methods outperform random and OMP baselines. Notably, C-SWAP removes up to 50\% of parameters without performance loss, surpassing all other methods.
For MobileNetV2 on ImageNet (fig. \ref{fig:short}.b), the model's high information density challenges OSP due to its compact design, resulting in steeper accuracy curves compared to other networks.

ResNet-50 results on ImageNet (fig. \ref{fig:short}.c) indicate that certain attribution methods are outperformed by OMP, except for C-BP and particularly C-SWAP, which consistently excel. This is attributed to the high class-specificity of some methods, making it difficult to score global neuron relevance in complex architectures and large datasets. In contrast, C-SWAP shows greater generality and effectively captures the global influence of neurons.
In ViT results (fig. \ref{fig:short}.d), explanation methods outperform baselines, with C-SWAP delivering the best results. Pruning ViT is particularly challenging due to its strong output interdependence inherent in its residuals, making it more complex than ResNet-50.

Overall, C-SWAP outperforms approaches shown in fig. \ref{fig:short}, highlighting its effective impact on model pruning. It demonstrates advantages over C-BP, which varies across seeds due to ranking stability issues, and AMP which is aggressive, fails to save essential structures.

\subsection{Assessing the strength of pruning criterion}
The benchmark results become more interpretable through the scalar values of SAUCE, which facilitate practical comparison between pruning criteria by assessing each method's ability to maintain model performance as pruning ratios increase.
Tab. \ref{tab:acc} presents SAUCE values for all models and datasets. C-SWAP consistently outperforms all baselines across all tasks, demonstrating a superior trade-off between model compression and performance.
While iLRP, Internal Influence, C-BP, and Conductance perform similarly on average, each exhibits unique strengths with specific architectures, highlighting the importance of selecting the appropriate attribution method for optimal pruning. Despite using a progressive pruning strategy like C-SWAP's, AMP proves unsuitable for pruning.
Due to implementation incompatibilities, results for iLRP on MobileNetV2, DeepLIFT on ViT are unavailable. 

\subsection{Neurons categories distribution}
\label{sec:distribution}
\begin{figure*}[t]
    \centering
    \begin{minipage}{0.45\textwidth}
    \centering
    \begin{table}[H]
        \centering
        \scriptsize
    \begin{tabular}{|l|c|c|c|}
    \hline
     & \textbf{Detrimental (\%)} & \textbf{Neutral (\%)} & \textbf{Critical (\%)} \\
    \hline
ResNet18 & $22.44{\scriptstyle\,\pm\,0.35}$ & $1.48{\scriptstyle\,\pm\,0.11}$ & $76.08{\scriptstyle\,\pm\,0.39}$ \\
ResNet50 & $7.14{\scriptstyle\,\pm\,0.08}$ & $27.37{\scriptstyle\,\pm\,0.37}$ & $65.49{\scriptstyle\,\pm\,0.38}$ \\
MobileNet & $12.29{\scriptstyle\,\pm\,0.21}$ & $8.35{\scriptstyle\,\pm\,0.56}$ & $79.36{\scriptstyle\,\pm\,0.68}$ \\
ViT & $23.18{\scriptstyle\,\pm\,0.13}$ & $32.27{\scriptstyle\,\pm\,0.14}$ & $44.55{\scriptstyle\,\pm\,0.27}$ \\\hline
    \end{tabular}
    \vspace{10mm} 
    \caption{Neuron distributions for each architecture.}
    \label{tab:neurons}
\end{table}
\end{minipage}%
    \hfill
    \begin{minipage}{0.45\textwidth}
        \centering
    \includegraphics[width=0.65\linewidth]{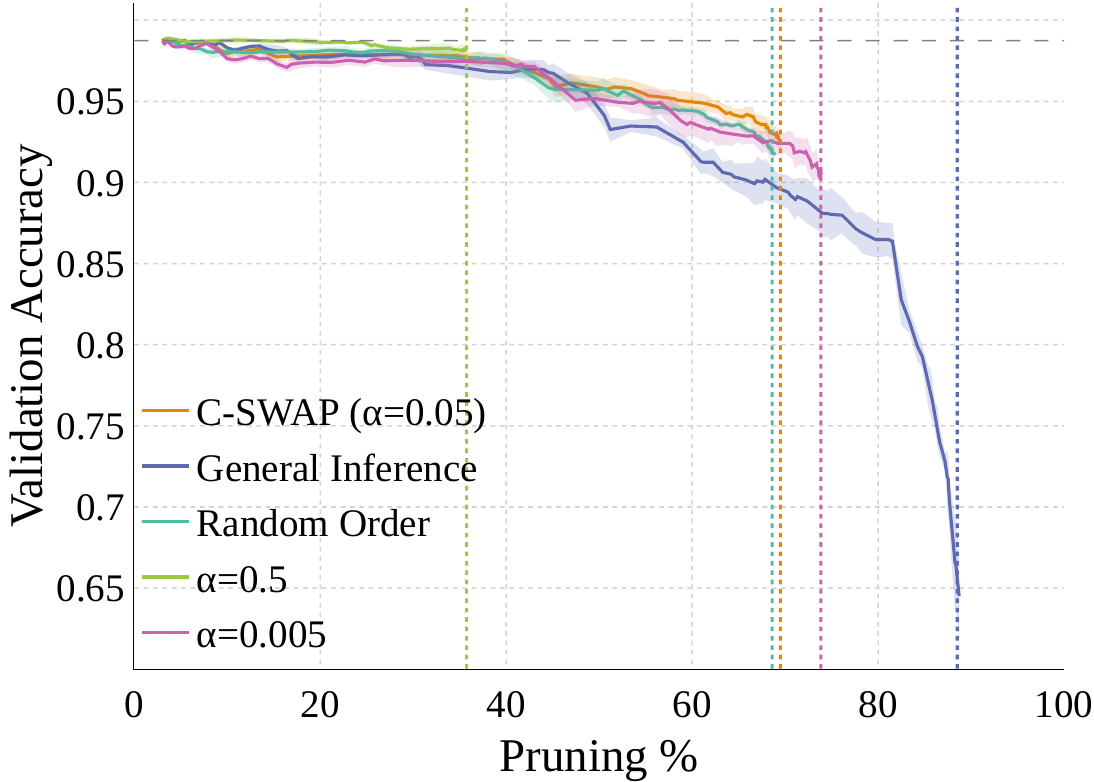}
    \caption{Ablation study removing only \textit{neutral} and \textit{detrimental} neurons.}
    \label{fig:ablation}
    \end{minipage}%
\end{figure*}

We examine the distribution of \textit{detrimental}, \textit{neutral}, and \textit{critical} neurons identified by our framework across the four architectures. Tab. \ref{tab:neurons} quantifies the overall proportion of each neuron category. For ResNet18, MobileNetV2, C-SWAP predominantly identifies neurons as critical. This outcome is attributed to the compact information encoded in these relatively small architectures. In contrast, the over-parameterization in ViT and ResNet50 leads to a higher proportion of neutral neurons, as their individual impact on the output is minimal. 
\subsection{Ablation study}
\label{sec:ablation}

We assess the impact of various factors in our methodology using ResNet-50. First, we evaluate the effectiveness of our voting strategy (def. \ref{def:crit}) by implementing a version of C-SWAP that naively compares distributions of scoring functions $\sigma_n$ and $\sigma_n^*$ across all classes, dubbed \textbf{general inference}. We also analyze the impact of different $\alpha$ values, the significance level for statistical testing. Finally, we conduct neuron permutation analysis to confirm that the order in which C-SWAP evaluates neuron importance within each layer does not significantly affect the results, using five seeds for permutations and reporting the average outcome.

Fig. \ref{fig:ablation} displays the results of these ablation studies on C-SWAP. We find that neuron analysis order has negligible impact on the pruning process. Optimizing the statistical test value ($\alpha$) proves valuable, as it governs the proportion of neurons identified as neutral. The general inference method, however, detects too many neurons as \textit{neutral} or \textit{detrimental}--false negatives--, resulting in overly aggressive pruning, and loss of critical information.

\subsection{Adaptation to semantic segmentation}

\label{sec:segmentation}
\begin{figure*}[t]
    \centering
    \begin{minipage}{0.49\textwidth}
    \begin{table}[H]
        \centering
        \footnotesize
    \begin{tabular}{l|c p{15pt} l|c}
    \cline{1-2}\cline{4-5}
     & \textbf{SAUCE} &  &  & \textbf{Percentage} \\
    \cline{1-2}\cline{4-5} 
Random & $10.3{\scriptstyle\,\pm\,1.2}$& & Detrimental  & $5.24{\scriptstyle\,\pm\,0.32}$ \\
AMP & $15.0{\scriptstyle\,\pm\,0.0}$   & & Neutral & $22.31{\scriptstyle\,\pm\,0.78}$ \\
OMP    & $17.9{\scriptstyle\,\pm\,0.0}$& & Critical & $72.45{\scriptstyle\,\pm\,0.57}$ \\
\cline{1-2}\cline{4-5}
\cline{1-2}\cline{4-5}
\textbf{C-SWAP} & $\mathbf{36.9{\scriptstyle\,\pm\,0.44}}$ & \multicolumn{1}{c}{} & \multicolumn{1}{c}{} & \\\cline{1-2} 
\multicolumn{2}{c}{\scriptsize{a) Comparative SAUCE scores}} & & \multicolumn{2}{c}{\scriptsize{b) C-SWAP neuron distributions}}
    \end{tabular}
    \caption{SAUCE scores (a) and neuron distributions (b) for DDRNet on cityscapes.}
    \label{tab:ddrnet}
\end{table}
\end{minipage}%
    \hfill
    \begin{minipage}{0.40\textwidth}
        \centering
 \includegraphics[width=0.75\linewidth]{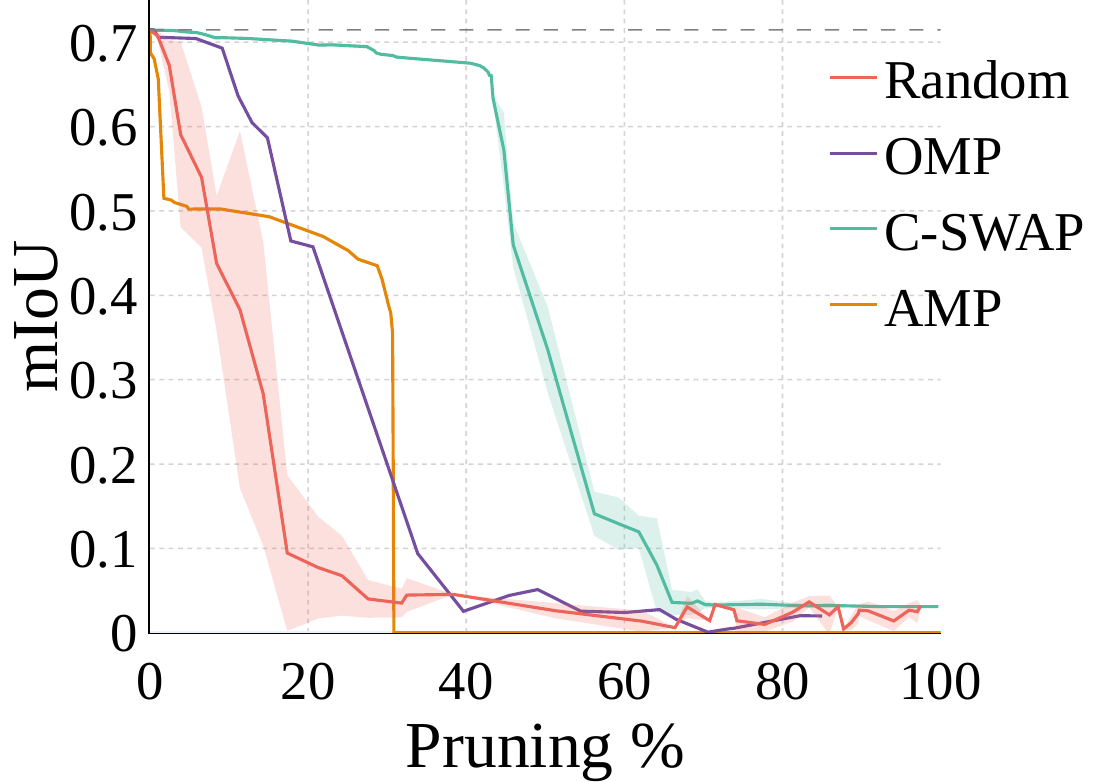}
    \caption{DDRNet on cityscapes.}
    \label{fig:ddrnet}
    \end{minipage}%
\end{figure*}

Interestingly, C-SWAP is readily extendable to more complex tasks such as semantic segmentation, by appropriately choosing $\sigma$. In our case, we consider the mean Intersection-over-Union (IoU) as $\sigma_{\text{Seg.}}$. 
To test our framework on segmentation models, we chose to analyze DDRNet 23S \cite{ddrnet} (6.3M parameters) pre-trained on cityscapes dataset \cite{cityscapes}, a widely used benchmark; the model's multi-branch, information-dense architecture provides a stringent testbed for C-SWAP. We restrict the segmentation comparison to Random, OMP, and AMP, as the other baselines are not applicable beyond classification. App. \ref{app:city} provides details regarding image selection in the segmentation setup.

As shown in fig. \ref{fig:ddrnet} and tab. \ref{tab:ddrnet}~a), C-SWAP maintains performance up to a 40\% pruning ratio; beyond this point, mIoU drops sharply as critical units (e.g., neurons/channels) are pruned; in contrast, other baselines begin degrading immediately. Only OMP removes a small portion of near-zero-weight units ($<20\%$) without noticeable impact. Tab. \ref{tab:ddrnet}~b) further indicates that DDRNet is already dense, with critical units predominating. Overall, these results demonstrate C-SWAP’s effectiveness for semantic segmentation, confirming its versatility beyond classification.

\section{Conclusion and Limitations}
\label{sec:conclusion}

We present C-SWAP, a one-shot structured pruning (OSP) algorithm that assigns causality-based relevance scores to model units (e.g., channels/filters, heads, blocks) and progressively prunes low-contribution structures. We show that attribution-guided criteria can outperform basic baselines for one-shot pruning, and C-SWAP consistently ranks best among baseline methods. Extensive evaluations, including with the newly proposed SAUCE metric, demonstrate superior Pareto trade-offs between model complexity and accuracy across CNN and ViT classifiers, without fine-tuning, and the approach extends to semantic segmentation.
These results underscore the critical role of explainable AI in advancing DNN compression.

\noindent\textbf{Limitations and future research work.}\,\,While our method sets a new benchmark for efficient structured pruning, it does have certain limitations. C-SWAP computes per-unit causal relevance, which is tractable for medium-width layers ($2048$ neurons $\sim$ 50min) but becomes computationally demanding for very wide layers (see app. \ref{app:time}). To mitigate this, exploring group-wise (block/channel) relevance scores is a promising research direction. 
Additionally, while our method shows promising results in pruning classification and semantic segmentation models, extending it to other complex tasks like object detection represents an exciting research avenue. Developing effective scoring functions for object-level tasks would provide valuable research opportunities for both explainability and neural network compression.

\bibliography{egbib}

\clearpage
\setcounter{page}{1}
\appendix

\section{Correlation Between Pruning Percentage And Compression}
\label{app:compression}

\begin{figure}[H]
\centering
        \includegraphics[width=0.5\linewidth]{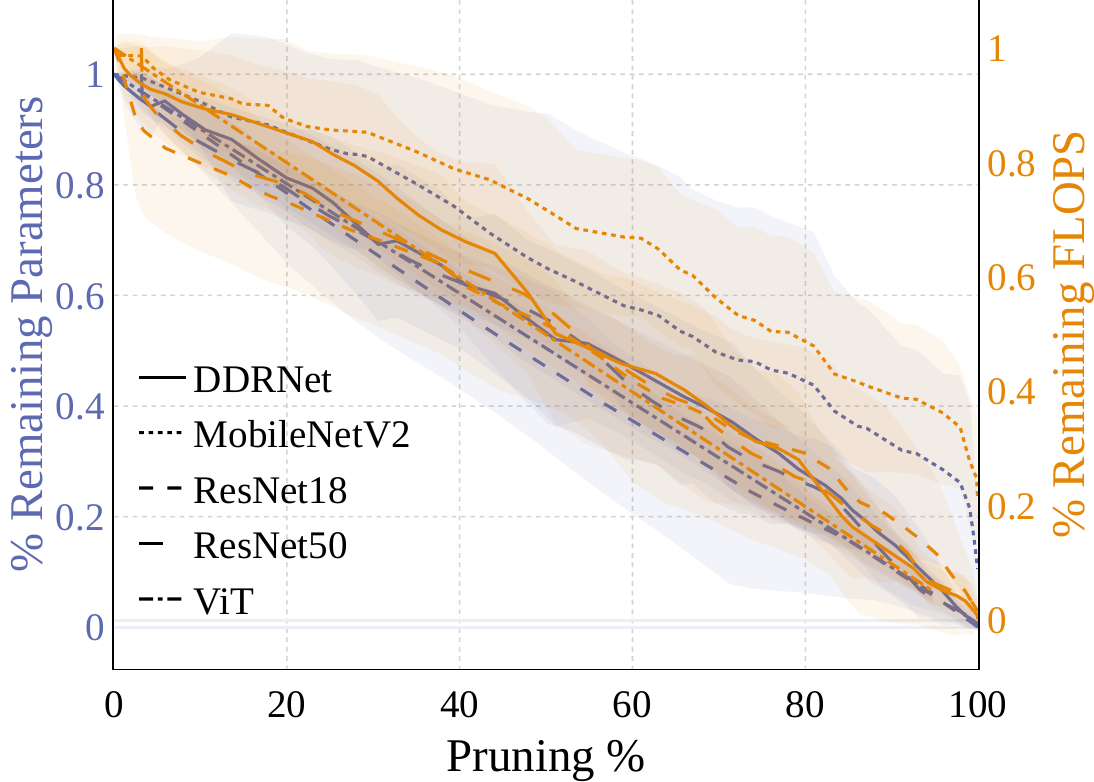}
  \caption{Correlation between pruning percentage, size reduction (number of parameters) and FLOPs reduction.}
  \label{fig:compression}
\end{figure}

In this section, we demonstrate that measuring the pruning percentage is an effective indicator of both memory gains and speed improvements achieved through pruning of the four networks under study. Fig \ref{fig:compression} illustrates that for all architectures, there is a near-linear correlation between speed-up (measured in FLOPs) and memory size (indicated by the number of parameters) with the pruning percentage. Conversely, for MobileNet, although a clear correlation exists, it is less linear compared to the other architectures due to its distinctive structure containing depth-wise convolution layers that are less amenable to pruning.

\section{Computational Costs}
\label{app:time}

\begin{table}[H]
    \centering
    \begin{tabular}{ll|c|c|c}
        Network & Dataset & Total Time (h:m:s) & Avg per Neuron & \# explored neurons \\ \hline
        ResNet18 & C10 & 1:58 & 1/17 s & 2 880\\
        MobileNetV2 & IN & 1:02:18 & 1/2.3 s & 9 128 \\
        DDRNet & Cit. $_{\text{\tiny{350 imgs}}}$ & 3:24:43 & 3.2 s & 3840\\
        ResNet50 & IN & 7:41:07 & 1.48 s & 19 008 \\
        ViT & IN & 34:41:08 & 2.70 s & 46 080 \\
    \end{tabular}
    \caption{Time consumption for a full run of C-SWAP on each analyzed network.}
    \label{tab:time}
\end{table}

Table~\ref{tab:time} presents the computational time required for a single run of C-SWAP on each network using \textbf{a single Nvidia 3090 GPU}, with 128 samples per class (or 350 samples in total in segmentation, see app. \ref{app:city}). Our results indicate that C-SWAP is highly efficient for compact architectures and small datasets. However, as network depth and layer width increase, the computational demand rises accordingly.

For structured pruning in ResNet and ViT architectures, the residual connections necessitate grouping residually-connected layers, which in turn accelerates the analysis process. This characteristic highlights an efficiency benefit inherent to networks with residual structures.

In the context of segmentation tasks, we observe that the computation of Intersection-over-Union (IoU) scores is more resource-intensive compared to the probability-based analysis used for classification. For example, DDRNet exhibits the longest average time per neuron, primarily because a new IoU score must be computed for each analyzed neuron. This suggests a potential direction for future work: identifying or designing more efficient metrics for segmentation.

In summary, while C-SWAP analysis can be time-consuming for architectures like ViT, these durations remain substantially lower than required for full model training and do not require post-analytical finetuning. This makes C-SWAP a pragmatic choice for performance evaluation despite its computational cost.

\section{Impact of Sample Size}
\label{app:samples}

\begin{figure}
    \centering
    \includegraphics[width=0.5\linewidth]{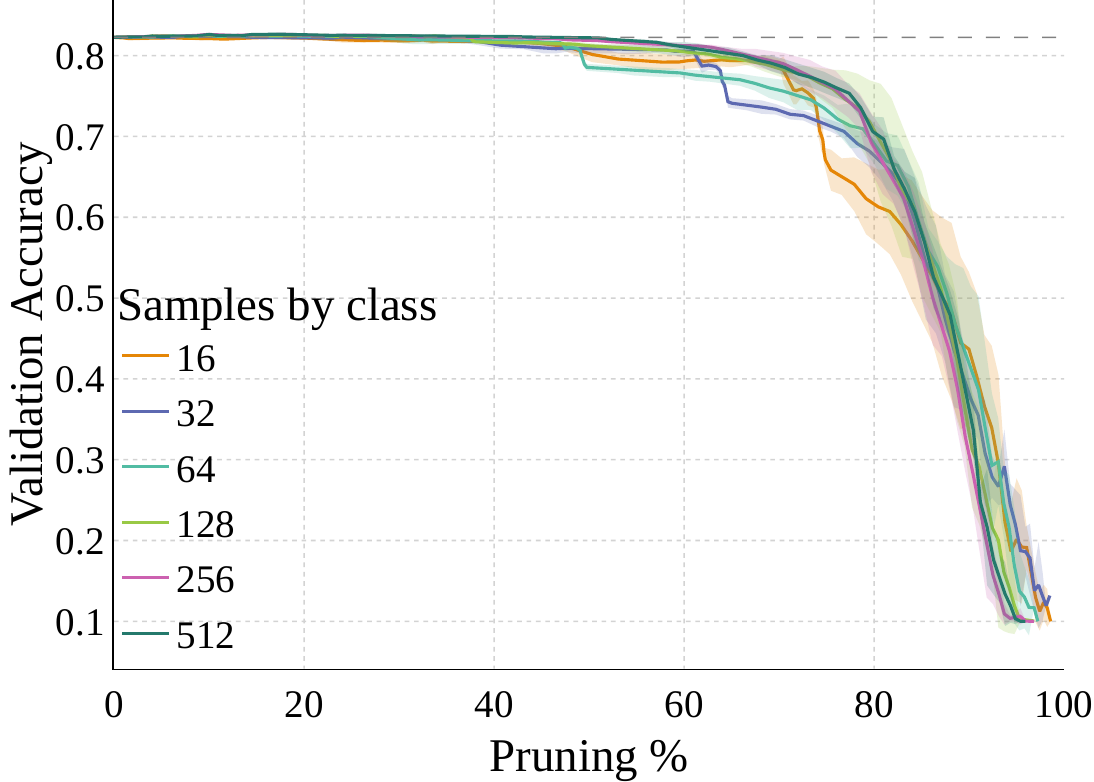}
    \caption{Effect of sample size $M$ on the pruning curve of C-SWAP for ResNet18 on CIFAR10. }
    \label{app:fig:samples}
\end{figure}

We examine how varying the sample size $M$ affects the performance of the C-SWAP algorithm on ResNet18 with CIFAR-10. As shown in fig. \ref{app:fig:samples}, C-SWAP consistently performs well over different $M$ values. Reducing $M$ slightly decreases pruning precision and performance, but the algorithm remains effective at identifying and removing less relevant neurons even with fewer samples. This robustness demonstrates C-SWAP’s efficiency and versatility with limited data. While higher $M$ can improve precision, C-SWAP is reliable and powerful regardless of sample size.

\section{ViT Structural Pruning}
\label{app:struct}
\begin{figure}
    \centering
    \includegraphics[width=0.3\linewidth]{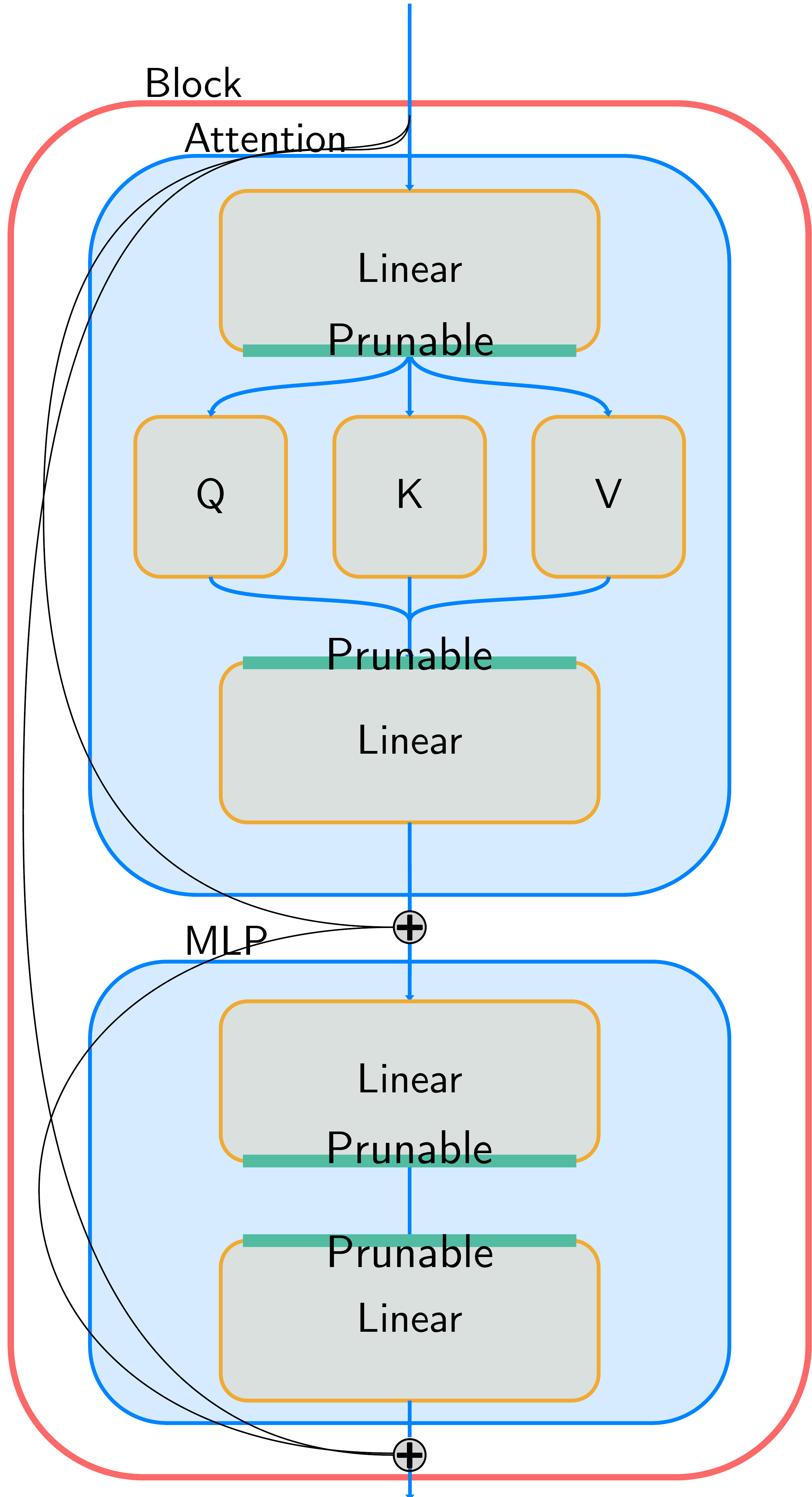}
    \caption{Schematized block of ViT. Highlighted in green, the prunable interfaces of ViT. All the others are linked by residual connections that are linked to token size.}
    \label{app:fig:vit}
\end{figure}

Structural pruning offers a practical approach to reducing model size by eliminating sub-networks without requiring additional compression overhead. This method is slightly constrained in most architectures due to the need to remove residually-connected neurons, and avoid the removal of crucial neurons such as those at the input and output (see fig. \ref{fig:pruning}). Nonetheless, transformers introduce a distinct challenge. Their highly-residual architecture exhibits significant interdependence between the input dimensions of layers and the token sizes, making structural pruning without intimate knowledge of the inner workings inherently complex.

In the case of ViT, understanding the specific architecture is critical due to the intricate connectivity between layers. Our study deliberately focused on pruning solely the independent dimensions: the inner dimension of the MLP layers (3072 neurons) and the attention layers (768x3 neurons), as schematized in fig. \ref{app:fig:vit}. This selective approach ensures that the essential token size dimensions remain intact, thereby preserving the functionality and performance of the model.
To substantiate this methodology, we examined the distribution of weights across the ViT architecture. Our findings revealed that a substantial portion of the weights are associated with either the prunable interfaces of the MLP or the attention layers. This significant proportion indicates that our targeted pruning approach has the potential to significantly reduce model size while maintaining the necessary structural and functional integrity.

Moreover, the challenge of pruning transformers like ViT lies in the balance between pruning efficiency and model performance. Since any reduction in token dimensions directly impacts the performance of the model, our strategy to concentrate on the independent dimensions proves to be pragmatic. This approach not only simplifies the pruning process but also ensures that the transformer retains its inherent advantages. 

\section{Additional Experimental Details}

\subsection{Sample selection for segmentation}
\label{app:city}

In the case of image classification, as discussed in the main paper, we selected 128 samples per class to ensure a representative subset of the data while avoiding the need to utilize the entire training set. However, for the semantic segmentation task, determining a fixed number of images per class is more challenging due to the varying frequency of class occurrence—some classes are present in all images, while others are considerably underrepresented.

To address this issue and to incorporate some degree of randomness in the sample selection process, we devised a two-step relaxed greedy image selection algorithm. In the first step, we greedily select images that contain objects from at least 15 different classes (out of the 19 classes in Cityscapes). We then gradually relax this constraint to ensure that there are at least 128 images containing each class. Rather than employing a fully greedy approach, our method intentionally introduces randomness, thereby promoting diversity in the selected subsets for different random seeds, while still guaranteeing coverage of all classes.

In the second step, we further filter the selected subset by removing images that contain only those classes which are already represented in more than 128 images, reducing the total number oof images required for the analysis, and potentially re-balancing the dataset. The full procedure is detailed in Alg~\ref{alg:filter_img}. As an example, for a specific seed (42), with 128 images per class, we selected a total of 353 images, with the majority class being present in 352 of them, and most classes being present in close to 128 images.

\SetKwComment{Comment}{\#\,}{}

\RestyleAlgo{ruled}
\SetAlgoNoEnd

\begin{algorithm}[t]
\small
\label{alg:filter_img}
\caption{Image filtering process for Cityscapes.}
\KwData{Cityscapes dataset: $\mathcal{I}$ ; Samples per class $n$}
$\{n^*_k\}_{k=1}^C \gets 0$\Comment*[r]{\scalebox{\algsize}{Number of samples in each class}}
$\mathcal{J} \gets \emptyset$\Comment*[r]{\scalebox{\algsize}{Subset of selected samples}}
$\Delta \gets 0$\Comment*[r]{\scalebox{\algsize}{Threshold decrease}}
\While{$\exists k \text{ s.t. } n^*_k < n$}{
\For{Random Image $I$ in  $\mathcal{I}$ }{
    $\mu_I = \sum_{k=1}^C\left[1\text{ if $k$ in $I$ and $n^*_k < n$ else $0$}\right]$ \Comment*[r]{\scalebox{\algsize}{Number of missing classes in the image}}
    \If{$\mu_I > 15-\Delta$}{$\mathcal{J} \gets \mathcal{J} \cup I$ \Comment*[r]{\scalebox{\algsize}{Add image to subset}}
    \For{$k$ present in $I$}{$n^*_k \gets n^*_k+1$}}}
    $\Delta \gets \Delta+1$\Comment*[r]{\scalebox{\algsize}{Decrease threshold}}}
\For{$I \in \mathcal{J}$}{
    $\nu_I = \prod_{k=1}^C\left[1\text{ if $k$ in $I$ and $n^*_k > n$ else $0$}\right]$ \Comment*[r]{\scalebox{\algsize}{1 if only majority classes in the image}}
    \If{$\nu_I = 1$ }{$\mathcal{J} \gets \mathcal{J} \setminus I$ \Comment*[r]{\scalebox{\algsize}{Remove image from subset}}}
  }
\KwResult{Image subset $\mathcal{J}$}

\end{algorithm}

\subsection{Adapted mechanistic pruning implementation}
\label{app:acdc}

\SetKwComment{Comment}{\#\,}{}

\RestyleAlgo{ruled}
\SetAlgoNoEnd

\begin{algorithm}[t]
\small
\label{alg:acdc}
\caption{Adapted Mechanistic Pruning Pseudo-Code.}
\KwData{Pre-trained NN: $F$ ; Samples $\{(x, y)\}$}
\KwData{Threshold: $\tau = 0.0575$}
$G \gets F$ 
\For{Layer $l$ in $[L-1, .., 1]$ (bottom up)}{
  \For{Neuron $n$ in Layer $l$ (sorted by magnitude)}{
    $\Tilde{G} \gets G$ cutting $\{n \rightarrow \nu \}$ for $\nu \in l+1$ \Comment*[r]{\scalebox{\algsize}{connections to neurons of next layer}}
    
  \If{$D_{KL}(F||\Tilde{G}) - D_{KL}(F||G) < \tau$}{
  $G \gets \Tilde{G}$ }}
  }
\KwResult{Pruned network $G$}

\end{algorithm}
Adapted Mechanistic Pruning  (AMP) is a baseline algorithm that we adapted from ACDC \cite{acdc}, an existing method focused on mechanistic interpretability. The original algorithm, aimed at mechanistic interpretability, utilizes $D_{KL}(F||\Tilde{G}) - D_{KL}(F||G)$, the difference between the KL divergences of the output distributions of the original model ($F$) and models with ($G$) and without ($\Tilde{G}$) the neuron to analyze. In our case, such a difference is computed  over all the classes of the multiclass task. While not originally intended for pruning, we straightforwardly adapted it to guide the pruning process in our implementation. 

AMP follows the progressive pruning strategy, where components are pruned based on the repurposed metric and a threshold value $\tau$ within the analysis process. This process allows for gradual reduction in model complexity while trying to maintain performance. The order in which we explore the layers and neurons is the same as in the progressive pruning strategy of C-SWAP: we explore first the layers that are the closest to the output of the model, and within the layers, we order the neurons by ascending magnitude of their weights. Our implementation of AMP is summarized in alg. \ref{alg:acdc}. It is as faithful as possible to the original method while being applicable to the problem at hand. We used the threshold value $\tau$ presented in the original paper \cite{acdc}, 0.0575.

\subsection{C-BP implementation}
\label{app:cbp}

\SetKwComment{Comment}{\#\,}{}

\RestyleAlgo{ruled}
\SetAlgoNoEnd

\begin{algorithm}[t]
\small
\label{alg:cbp}
\caption{C-BP algorithm.}
\KwData{Pre-trained NN: $F$ ; Samples $\{(x, y)\}$}
\KwData{Scoring function: $\sigma(x)$; Threshold: $\alpha = 0.05$}
\For{Layer $l$ in $[L-1, .., 1]$ (bottom up)}{
  \For{Neuron $n$ in Layer $l$ (sorted by magnitude)}{
    $\Tilde{F} \gets F$ cutting $\{n \rightarrow \nu \}$ for $\nu \in \mathcal{C}_{|l+1}$ \Comment*[r]{\scalebox{\algsize}{connections to critical neurons of next layer}}
    \For{Class $k$ in $\{1, C\}$}{
        Compute $\sigma_{n, i, k}$ \Comment*[r]{\scalebox{\algsize}{perturbed scores}}
      Compute $\pi_n^{(k)}$ \Comment*[r]{\scalebox{\algsize}{class inference predicate}}
    }
  Compute $\xi_n$  \Comment*[r]{\scalebox{\algsize}{general causal effect}}
  \If{$\logor{k=1}{C} \pi_n^{(k)}$}{
  \If{$\xi_n \leq 0$}{$\mathcal{C} \gets \mathcal{C} \cup \{n\} $ \Comment*[r]{\scalebox{\algsize}{neuron is critical}} }
  \If{$\xi_n > 0$}{$\mathcal{N} \gets \mathcal{N} \cup \{n\} $  \Comment*[r]{\scalebox{\algsize}{neuron is detrimental}} }
  }
  \Else{$\mathcal{N}e \gets \mathcal{N}e  \cup \{n\} $  \Comment*[r]{\scalebox{\algsize}{neuron is neutral} }}
  }
}
\For{Neuron $n$ in $\mathcal{N}$ (ranked by $\xi_n$)}{Prune $n$ from $F$ \Comment*[r]{\scalebox{\algsize}{Remove detrimental neurons ranked by CE}}}
\For{Neuron $n$ in $\mathcal{N}e$ (ranked by $|\xi_n|$)}{Prune $n$ from $F$ \Comment*[r]{\scalebox{\algsize}{Remove neutral neurons ranked by absolute CE}}}
\For{Neuron $n$ in $\mathcal{C}$ (ranked by $\xi_n$)}{Prune $n$ from $F$ \Comment*[r]{\scalebox{\algsize}{Remove critical neurons ranked by CE}}}
\KwResult{Pruned network $F$}

\end{algorithm}

To implement C-BP, we followed the conventional "ranking then pruning" strategy used by the other XAI baselines. To do so, we implemented the C-BP algorithm as summarized in alg. \ref{alg:cbp}. It follows the same principles as other attribution-based methods: first it ranks the neurons, with the particularity of separating them in Detrimental, Neutral Critical categories, and then it prunes the neurons based on the ranking. 

\section{Higher Scale Figures}
\label{app:reso}
\begin{figure}
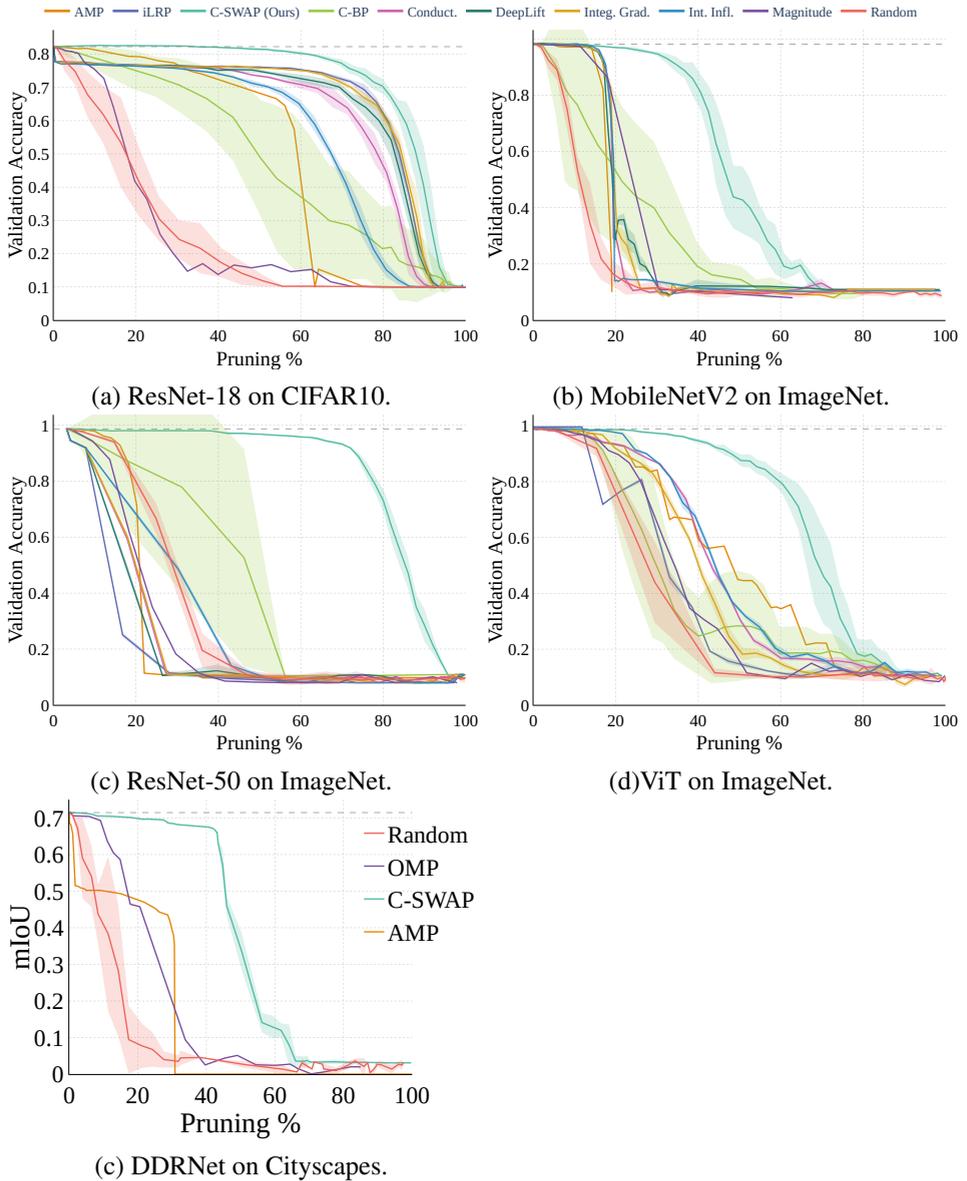

\newcommand{\mysize}{0.485}

\begin{tabular}{cc}
 \multicolumn{2}{c}{\includegraphics[width=0.95\linewidth, clip, trim=0.2cm 0.4cm 0.2cm 1.85cm,]{legend.pdf}}\\
\includegraphics[width=\mysize\linewidth]{ResNet18_CIFAR10_acc.pdf}&
\includegraphics[width=\mysize\linewidth]{MobileNet_MiniImageNet_acc.pdf}\\
(a) ResNet-18 on CIFAR10.&(b) MobileNetV2 on ImageNet.\\
\includegraphics[width=\mysize\linewidth]{ResNet50_MiniImageNet_acc.pdf}&
\includegraphics[width=\mysize\linewidth]{ViT_MiniImageNet_acc.pdf}\\
(c) ResNet-50 on ImageNet.&(d)ViT on ImageNet.\\
\includegraphics[width=\mysize\linewidth]{res_seg_bmvc.pdf}&\\
(c) DDRNet on Cityscapes.&
\end{tabular}   
\caption{Validation accuracy as a function of percentage of parameters removed for three architectures. Dashed line represents original model performance. }
\label{app:fig:short}
\end{figure}

\begin{figure}
\newcommand{\mysize}{0.495}

\begin{tabular}{cc}
\includegraphics[width=\mysize\linewidth]{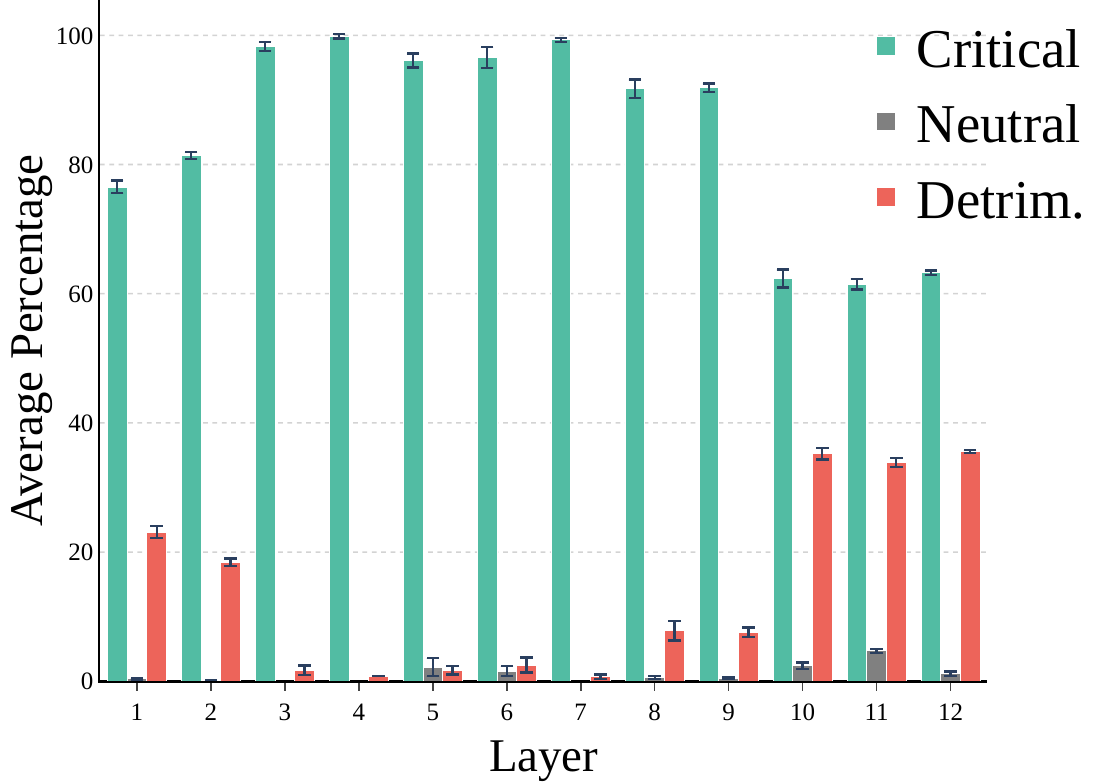}&
\includegraphics[width=\mysize\linewidth]{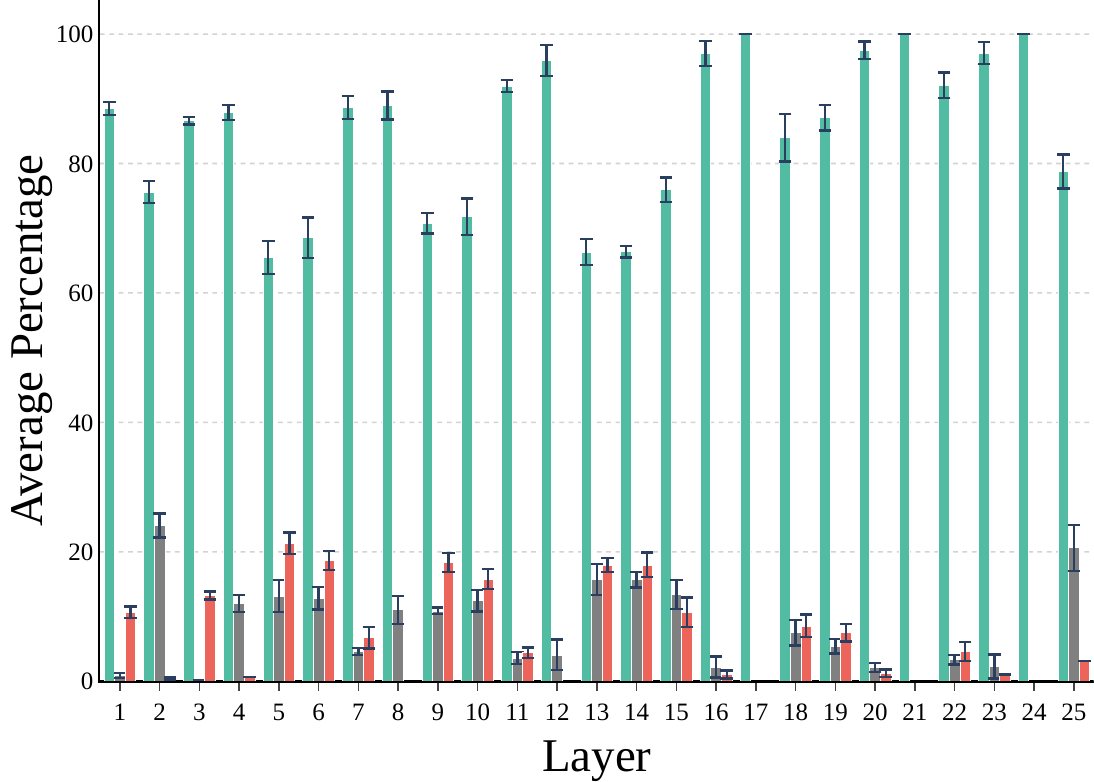}\\
(a) ResNet-18 &(b) MobileNetV2 \\
\includegraphics[width=\mysize\linewidth]{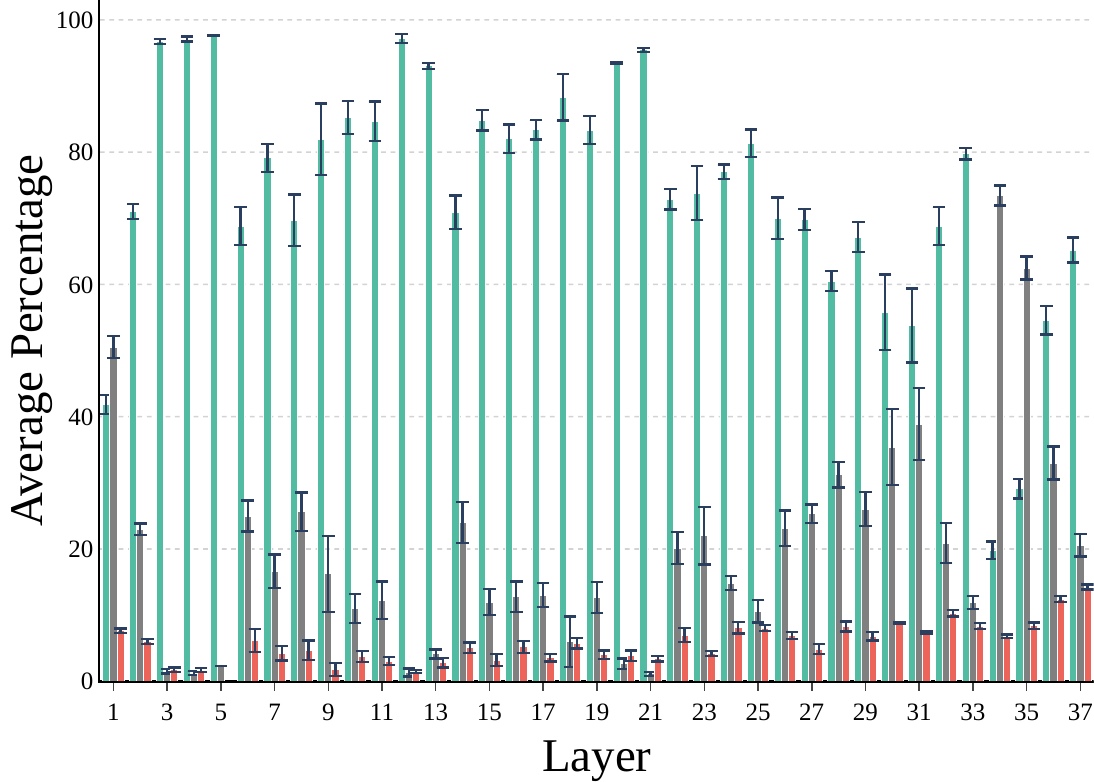}&
\includegraphics[width=\mysize\linewidth]{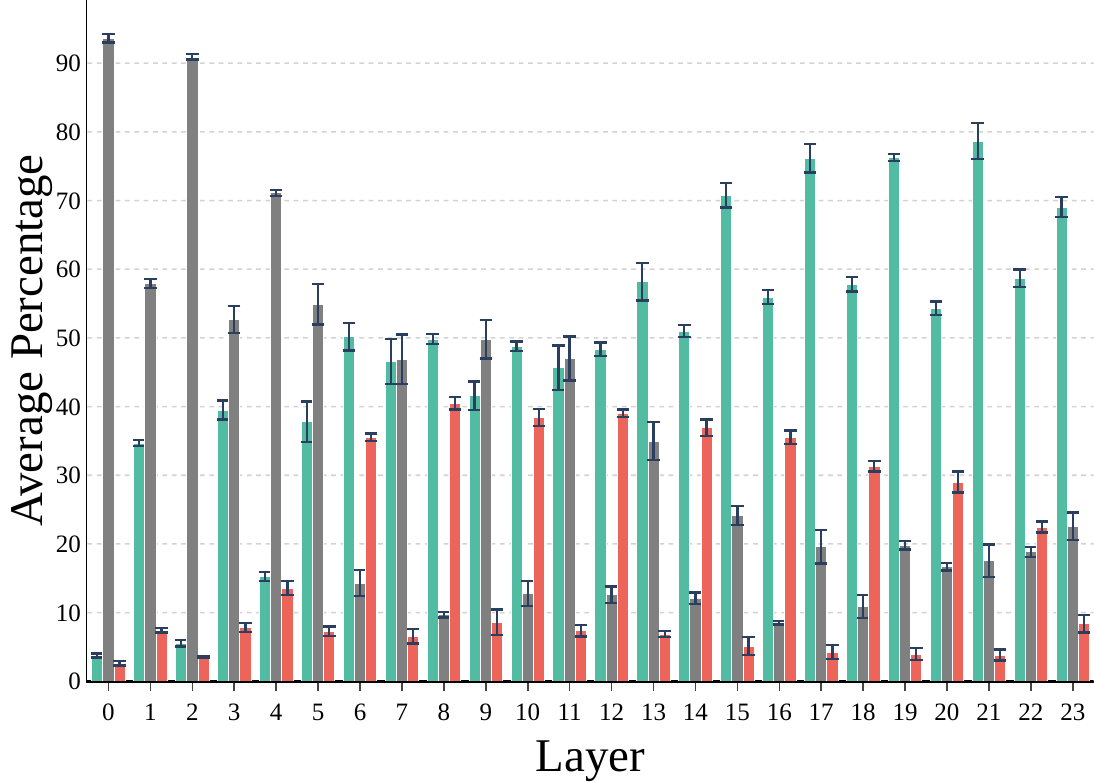}\\
(c) ResNet-50 &(d) ViT
\end{tabular}
\caption{Neuron type (\%) for each layer of the models. Green bars are critical neurons, greys, neutral ones and reds, detrimental ones.}
\label{app:fig:neur}
\end{figure}

In this section, we provide the figures in the paper on a larger scale. See figs. \ref{app:fig:short} and \ref{app:fig:neur}.

\end{document}

%% file: mlp_res.tex
  \definecolor{amber}{HTML}{f16767}
    \scalebox{0.70}{\begin{tikzpicture}[node distance=10.7cm and 10.8cm, align=center]
\node[circle, draw, minimum size=0.4cm] (I1) at (0,1) {1};
\node[circle, draw, minimum size=0.4cm] (I2) at (0,0) {2};
\node[circle, draw, minimum size=0.4cm] (I3) at (0,-1) {3};
\node[circle, draw, ultra thick, amber, minimum size=0.4cm] (H11) at (1.5,1) {1};
\draw (1.5,1) node[cross,] {};
\node[circle, draw, minimum size=0.4cm] (H12) at (1.5,0) {2};
\node[circle, draw, minimum size=0.4cm] (H13) at (1.5,-1) {3};
\node[circle, draw, minimum size=0.4cm] (H21) at (3,0.5) {1};
\node[circle, draw, minimum size=0.4cm] (H22) at (3,-0.5) {2};
\node[circle, draw, ultra thick, amber,  minimum size=0.4cm] (O1) at (4.5,1) {1};
\node[circle, draw, minimum size=0.4cm] (O2) at (4.5,0) {2};
\node[circle, draw, minimum size=0.4cm] (O3) at (4.5,-1) {3};
\node[circle, draw, minimum size=0.4cm] (k1) at (6,1) {1};
\node[circle, draw, minimum size=0.4cm] (k2) at (6,0) {2};
\node[circle, draw, minimum size=0.4cm] (k3) at (6,-1) {3};
\draw[->, ultra thick, amber] (I1) -- (H11);
\draw[->, ultra thick, amber] (I2) -- (H11);
\draw[->, ultra thick, amber] (I3) -- (H11);
\draw[->] (I1) -- (H12);
\draw[->] (I2) -- (H12);
\draw[->] (I3) -- (H12);
\draw[->] (I1) -- (H13);
\draw[->] (I2) -- (H13);
\draw[->] (I3) -- (H13);
\draw[->, ultra thick, amber] (H11) -- (H21);
\draw[->] (H12) -- (H21);
\draw[->] (H13) -- (H21);
\draw[->, ultra thick, amber] (H11) -- (H22);
\draw[->] (H12) -- (H22);
\draw[->] (H13) -- (H22);
\draw[->, ultra thick, amber] (H21) -- (O1);
\draw[->, ultra thick, amber] (H22) -- (O1);
\draw[->] (H21) -- (O2);
\draw[->] (H22) -- (O2);
\draw[->] (H21) -- (O3);
\draw[->] (H22) -- (O3);
\draw[->, ultra thick, amber] (O1) -- (k1); 
\draw[->] (O2) -- (k1); 
\draw[->] (O3) -- (k1); 
\draw[->, ultra thick, amber] (O1) -- (k2); 
\draw[->] (O2) -- (k2); 
\draw[->] (O3) -- (k2); 
\draw[->, ultra thick, amber] (O1) -- (k3); 
\draw[->] (O2) -- (k3); 
\draw[->] (O3) -- (k3); 
\foreach \i in {2,3} {
    \draw[->, dotted] (H1\i) -- (O\i);
}
\draw[->, ultra thick, dotted, amber] (H11) -- (O1);
\node[above=0.3cm] at (I1) {Layer 1};
\node[above=0.3cm] at (H11) {Layer 2};
\node[above=0.8cm] at (H21) {Layer 3};
\node[above=0.3cm] at (O1) {Layer 4};
\node[above=0.3cm] at (k1) {Layer 5};
\end{tikzpicture}}